\documentclass{article}

\PassOptionsToPackage{table,xcdraw}{xcolor}
\PassOptionsToPackage{authoryear}{natbib} 

\newcommand{\DiNo}{\textit{DiNo}}
\newcommand{\RanBu}{\textit{RanBu}}

\usepackage{tikz}

\tikzset{
  treenode/.style = {align=center, inner sep=0pt, text centered,
    font=\sffamily},
  arn_n/.style = {treenode, circle, white, font=\sffamily\bfseries, draw=black,
    fill=black, text width=1.5em},
  arn_r/.style = {treenode, circle, red, draw=red, 
    text width=1.5em, very thick},
  arn_x/.style = {treenode, rectangle, draw=black,
    minimum width=0.5em, minimum height=0.5em}
}

\usepackage{arxiv}

\usepackage{makecell}
\usepackage[utf8]{inputenc} 
\usepackage[T1]{fontenc}    
\usepackage{url}            
\usepackage{booktabs}       
\usepackage{amsfonts}       
\usepackage{amssymb}        
\usepackage{nicefrac}       
\usepackage{microtype}      
\usepackage{lipsum}		
\usepackage{graphicx}
\usepackage{doi}
\usepackage{amsmath}
\usepackage{float}

\usepackage{multirow}
\usepackage{colortbl}             
\usepackage[table,xcdraw]{xcolor}

\usepackage{enumitem}
\usetikzlibrary{positioning, calc}
\usepackage[authoryear]{natbib}

\usepackage[ruled,vlined]{algorithm2e}

\usepackage{rotating}

\usepackage{amsthm}
\usepackage{hyperref}       

\theoremstyle{definition}

\title{DiNo and RanBu: Lightweight Predictions from Shallow Random Forests}

\author{ \href{https://orcid.org/0000-0002-4223-9162}{\includegraphics[scale=0.06]{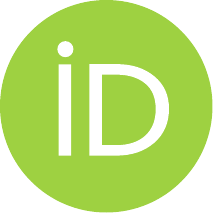}\hspace{1mm}Tiago Mendon\c{c}a dos Santos}\\
        Insper\\ Institute of Education and Research\\ São Paulo, Brazil \\
	\texttt{tiagoms1@insper.edu.br} \\
        \And
        \href{https://orcid.org/0000-0003-0379-9690}{\includegraphics[scale=0.06]{orcid.pdf}\hspace{1mm}Rafael Izbicki}\\
        Department of Statistics\\ Federal University of São Carlos\\ São Paulo, Brazil\\
        \texttt{rizbicki@ufscar.br}\\
        \And
        Lu\'{i}s Gustavo Esteves\\
        Institute of Mathematics and Statistics\\ University of São Paulo\\ São Paulo, Brazil\\
        \texttt{lesteves@ime.usp.br}
}

\date{}

\renewcommand{\headeright}{}
\renewcommand{\undertitle}{}
\renewcommand{\shorttitle}{}

\hypersetup{
pdftitle={DiNo and RanBu: Lightweight Predictions from Shallow Random Forests},
pdfauthor={Tiago ~Mendon\c{c}a dos Santos, Rafael ~Izbicki, Lu\'{i}s Gustavo ~Esteves},
pdfkeywords={Random Forest, Tree Proximity, Cophenetic Distance, Kernel Methods, Shallow Forests, Quantile Regression, Fast Inference, Tabular Data},
}

\begin{document}
\maketitle

\begin{abstract}

Random Forest ensembles are a strong baseline for tabular prediction tasks, but their reliance on hundreds of deep trees often results in high inference latency and memory demands, limiting deployment in latency-sensitive or resource-constrained environments. We introduce \DiNo{} (Distance with Nodes) and \RanBu{} (Random Bushes), two shallow-forest methods that convert a small set of depth-limited trees into efficient, distance–weighted predictors. \DiNo{} measures cophenetic distances via the most recent common ancestor of observation pairs, while \RanBu{} applies kernel smoothing to Breiman’s classical proximity measure. Both approaches operate entirely after forest training: no additional trees are grown, and tuning of the single bandwidth parameter $h$ requires only lightweight matrix–vector operations. Across three synthetic benchmarks and 25 public datasets, \RanBu{} matches or exceeds the accuracy of full-depth random forests—particularly in high-noise settings—while reducing training plus inference time by up to 95\%. \DiNo{} achieves the best bias–variance trade-off in low-noise regimes at a modest computational cost. Both methods extend directly to quantile regression, maintaining accuracy with substantial speed gains. The implementation is available as an open-source R/C++ package at \url{https://github.com/tiagomendonca/dirf}. We focus on structured tabular random samples (i.i.d.), leaving extensions to other modalities for future work.

\end{abstract}


\section{Introduction}
\label{sec:intro}

Tree ensembles—most notably \emph{Random Forests} (RF; \citealt{breiman2001rf})—are a mainstay in tabular machine–learning pipelines. They handle heterogeneous features without special preprocessing, are robust to over-fitting, and require minimal hyper-parameter tuning \citep{breiman2001rf,BiauScornet2016}. Yet, in practical deployments where models must respond quickly and operate under hardware constraints, two bottlenecks often emerge when RFs grow to hundreds of deep trees:

\begin{enumerate}[label=(\roman*)]
\item \textbf{Inference latency.} Traversing many deep trees per query incurs irregular memory access patterns, dominating CPU time and straining cache performance—an obstacle for low-latency applications such as user-facing APIs, streaming, or embedded devices.

\item \textbf{Sensitivity to irrelevant features.} In high-dimensional or noisy settings, splits on uninformative variables degrade both statistical efficiency and computational performance, slowing split selection and increasing model size.
\end{enumerate}

Research on accelerating ensemble inference ranges from hardware-aware optimizations \citep{browneforestpacking, romero2022bolt} to algorithmic compression and shallow-tree strategies \citep{kumar2017bonsai}. While effective, these solutions often require specialized engineering or retraining.

\paragraph{Our approach.}
We show that \emph{shallow} forests—tens of trees with modest, fixed depth—retain enough geometric structure to support competitive predictors when transformed into a \emph{kernel-like weighting scheme}. We propose creating such weights using two schemes:

\begin{description}
    \item[\textbf{DiNo}]  
    A cophenetic distance (\citet{cofenetica}) that, for each tree, counts the number of edges from each of two observations to their \emph{most recent common ancestor} (MRCA) and takes the maximum. A Gaussian kernel smoothing weighting of the forest-averaged distance yields smooth, distance-weighted predictions.

    \item[\textbf{RanBu}]  
    A Gaussian kernel-like reweighting of Breiman’s \emph{distance}—one minus the average proportion of trees in which two points share the same leaf. This continuous measure in $[0,1]$ emphasizes closer pairs while smoothly attenuating more distant ones.
\end{description}

Both operate entirely \emph{post hoc}: the forest structure is fixed, and tuning the single bandwidth parameter~$h$ involves only lightweight matrix–vector operations over precomputed leaf responses. Shallow trees make inference efficient, enabling deployment in large-scale or interactive settings.

The MRCA distance captures hierarchical separation via the longest branch from each leaf to their MRCA, making it less sensitive to split balance. Gaussian rescaling turns both into smooth, locality-aware similarities. As they require no retraining, these measures enable post hoc enhancement of shallow ensembles with minimal computational cost.

\vspace{0.5\baselineskip}
\noindent
\textbf{Illustrative speed-up.}  
On the \textsc{Online News Popularity} dataset (39k rows, 59 predictors), a 500-tree full-depth RF requires 95.6 seconds to train and predict for one test point on a machine with an Intel Core i7-11800H processor, 64 GB of RAM, running Windows 11. \RanBu{}, with 50 trees of depth~5—the same as \DiNo{}—takes 1.26 seconds for both steps, a nearly 75-fold reduction in total computation time with negligible change in test MSE.

\subsection*{Contributions.}
\vspace{-0.3\baselineskip}
Our contributions are the following:
\begin{enumerate}[label=(\arabic*)]
\item \textbf{Tree–based kernels.}  
      We develop two tree-based smoothing kernels: the MRCA distance and Breiman's distance. We show  that a Gaussian rescaling of these distances obtained in random forests yields effective kernel smoothers for prediction.
\item \textbf{Lightweight estimators.}  
      We introduce \DiNo{} and \RanBu{} for mean and quantile regression, designed for shallow, fixed-depth forests with only post-training bandwidth tuning.
\item \textbf{Empirical evaluation.}  
   We evaluate \RanBu{} and \DiNo{} on three synthetic and 25 public datasets. \RanBu{} matches or surpasses RF in accuracy—especially under high noise—while reducing inference time by up to 95\%. \DiNo{} achieves the best bias–variance trade-off in low-noise settings. On real-world datasets, \RanBu{} consistently outperforms GRF in predictive accuracy and runtime, and secures advantages over Boosting and Reduced R.F., whereas \DiNo{} delivers systematic improvements over the Reduced R.F.\ baseline, confirming that the kernel weighting step adds predictive value beyond simple pruning.
\item \textbf{Open-source implementation.}  
        We release a production-ready \textsf{R}/C++ implementation at \url{https://github.com/tiagomendonca/dirf}, enabling reproducibility and practical use.
\end{enumerate}

\subsection*{Paper outline}
Section~\ref{sec:related} reviews related work on proximity measures and efficient RF variants. Section~\ref{sec:method_shallow} details the proposed distances and predictors. Section~\ref{sec:experiments} presents synthetic and real-data experiments. Section~\ref{sec:applications} discusses real-world implications. Section~\ref{sec:conclusion} summarises contributions and outlines future directions.

\section{Related Work}
\label{sec:related}

\subsection{Tree-Based Proximity Measures and Shallow Forests}

Random Forests (RF) naturally induce a notion of similarity between instances based on how often they fall in the same terminal node, as introduced by \citet{breiman2004rftools} and \citet{Breiman_proximity}. This co-occurrence-based proximity has been widely used in unsupervised tasks such as clustering and outlier detection, but its binary nature—treating all co-occurring pairs equally—limits expressiveness. To address this, several works have enriched proximities with structural information from the trees. For example, \citet{englund2012proximity} proposed a depth-aware measure that decays with the distance between terminal nodes, while \citet{panda2024learninginterpretablecharacteristickernels} demonstrated that the RF proximity matrix is positive-definite and characteristic, reinforcing its potential for kernel-based learning.

\DiNo{} and \RanBu{} build on this tradition with two innovations aimed at balancing predictive accuracy and computational efficiency. First, both methods operate on a deliberately small number of shallow trees, reducing inference latency without altering the standard RF training pipeline. Second, they employ continuous, task-adapted similarity functions: \DiNo{} derives a distance from the most recent common ancestor (MRCA) in each tree, while \RanBu{} applies a Gaussian kernel smoothing weighting to the classical RF proximity. This design preserves compatibility with existing RF tooling and interpretability of axis-aligned splits, while enabling smooth weighting of training observations at prediction time. The result is a family of lightweight estimators particularly suited for latency-sensitive scenarios, requiring only post-training computations over precomputed leaf responses.

\subsection{Efficient Inference via Compression and Hardware Acceleration}

System-level optimizations have also been proposed to accelerate ensemble inference. Forest Packing \citep{browneforestpacking} improves cache efficiency by reorganizing tree nodes in memory. Bolt \citep{romero2022bolt} precomputes decision paths and stores them in compressed lookup tables, while GPU-based methods such as Tahoe \citep{xieTahoe} exploit parallelism for further gains. These approaches can yield substantial speed-ups but often depend on specialized hardware or significant engineering effort. Overall, prior work on efficient tree-based learning has largely followed two main directions: (1) compressing or simplifying models to reduce evaluation time, and (2) optimizing inference pipelines through specialized software or hardware techniques.

Unlike these methods, \DiNo{} and \RanBu{} achieve efficiency entirely at the algorithmic level, without hardware-specific requirements or memory layout modifications. By operating with a deliberately compact configuration —typically 50 trees with a maximum depth of 5 — and reusing precomputed leaf assignments for fast distance computation, they offer a simple and portable alternative that can be seamlessly integrated into a wide range of production environments.

Our methods build on the idea that a depth-limited random forest captures rich structural information about the training data. We extract this information in the form of tree-based distances and transform them into continuous similarity weights via a Gaussian kernel.

\section{Shallow-Forest Kernels and Proposed Methods}
\label{sec:method_shallow}

Our methods build on the idea that a depth-limited random forest can provide rich structural information about the training data. We extract this information in the form of tree-based distances and transform them into continuous similarity weights via a Gaussian kernel smoothing weighting scheme. Two variants are considered: \DiNo{}, based on a most-recent common ancestor (MRCA) distance, and \RanBu{}, based on the classical Breiman proximity. Both can be extended from conditional mean estimation to conditional quantile estimation through a distance-weighted empirical distribution function.

Section~\ref{ssec:dists} defines the two distances; Section~\ref{ssec:kernelschemes} describe the mean estimators; and Section~\ref{ssec:qr} presents the quantile regression extension.

\subsection{Forest-based distances}
\label{ssec:dists}

Let $\mathcal{F} = \{T_b\}_{b=1}^B$ be a random forest of $B$ trees, each with maximum depth $d_{\max}$. In our experiments, we typically set $B = 50$ and $d_{\max} = 5$ to limit computational cost while preserving sufficient structural information. Denote by $\ell(i,b)$ the terminal node index for observation $i$ in tree $T_b$, and by $\ell(\mathbf{x},b)$ the terminal node of a query $\mathbf{x}$. We use $h>0$ for the bandwidth parameter and $\mathbb{I}$ for the indicator function.

\paragraph{MRCA distance.}
For two terminal nodes $n_1$ and $n_2$ in a binary tree $T$, let $\mathrm{mrca}(n_1,n_2)$ be their most recent common ancestor. For any two nodes $a,b \in T$, denote by $\#(a,b)$ the number of edges on the unique path between $a$ and $b$. The MRCA distance is
\[
d_T^{\mathrm{mrca}}(n_1,n_2) =
\max\!\Bigl\{\#\bigl(n_1,\operatorname{mrca}(n_1,n_2)\bigr),\;
\#\bigl(n_2,\operatorname{mrca}(n_1,n_2)\bigr)\Bigr\}.
\]
By taking the maximum, the distance is determined by the longer of the two branches from the MRCA to the leaves, capturing the dominant component of their separation and remaining stable under minor depth imbalances.

The forest-level MRCA distance is the average over trees:
\[
d^{\mathcal{F}}(\mathbf{x},\mathbf{x}') 
=
\frac{1}{B}\sum_{b=1}^B d_{T_b}^{\mathrm{mrca}}\bigl(\ell(\mathbf{x},b), \ell(\mathbf{x}',b)\bigr),
\]
rescaled to $[0,1]$. A formal proof that this function satisfies the properties of a distance metric is provided in the Appendix.

\paragraph{Breiman distance.}
The Breiman distance \citep{breiman2004rftools} is
\[
d^{\mathrm{br}}(\mathbf{x},\mathbf{x}')
=
1 - \frac{1}{B} \sum_{b=1}^{B} \mathbb{I}\bigl\{\ell(\mathbf{x},b) = \ell(\mathbf{x}',b)\bigr\},
\]
the complement of the classical RF proximity. Unlike the binary nature of the proximity in each tree, its forest-level average is continuous in $[0,1]$, similar to the MRCA distance.

\subsection{\DiNo{} and \RanBu{}: proposed kernel-like weighting schemes}
\label{ssec:kernelschemes}

The proposed methods, \DiNo{} and \RanBu{}, transform discrete tree-based distances into smooth, continuous weights via a Gaussian kernel smoothing weighting scheme. Given a query $\mathbf{x}$, the generic weight for observation $i$ is
\[
w_i(\mathbf{x}) =
\frac{\exp\bigl[-(d(\mathbf{x},\mathbf{X}_i)/h)^2\bigr]}
{\sum_{j=1}^n \exp\bigl[-(d(\mathbf{x},\mathbf{X}_j)/h)^2\bigr]},
\]
where $d(\cdot,\cdot)$ is a forest-based distance and $h>0$ is a bandwidth parameter controlling locality.

\paragraph{\DiNo{}.} Uses the MRCA-based distance $d^{\mathcal{F}}$ from Section~\ref{ssec:dists}, which captures the longest branch from the most recent common ancestor in shallow trees, offering robustness to unbalanced splits. The conditional mean estimator is

\begin{equation}
\hat r_{\mathrm{DiNo}}(\mathbf{x})
=
\sum_{i=1}^n 
\frac{\exp\bigl[-(d^{\mathcal{F}}(\mathbf{x},\mathbf{X}_i)/h)^2\bigr]}
{\sum_{j=1}^n \exp\bigl[-(d^{\mathcal{F}}(\mathbf{x},\mathbf{X}_j)/h)^2\bigr]}
\, Y_i.
\label{formula:dino}
\end{equation}

\paragraph{\RanBu{}.} Replaces $d^{\mathcal{F}}$ with the Breiman distance $d^{\mathrm{br}}$, the complement of the classical RF proximity:
\begin{equation}
\hat r_{\mathrm{RanBu}}(\mathbf{x})
=
\sum_{i=1}^n 
\frac{\exp\bigl[-(d^{\mathrm{br}}(\mathbf{x},\mathbf{X}_i)/h)^2\bigr]}
{\sum_{j=1}^n \exp\bigl[-(d^{\mathrm{br}}(\mathbf{x},\mathbf{X}_j)/h)^2\bigr]}
\, Y_i.
\label{formula:ranbu}
\end{equation}

\textbf{Computation Shortcut.} At first sight, computing Equations~\ref{formula:dino} and~\ref{formula:ranbu} may appear computationally demanding, since one would need to store the entire training set $X_1, \dots, X_n$ and evaluate all pairwise distances. In practice, however, it suffices to store only the leaf assignments of each observation. For instance, with $n = 1{,}000$ training samples and $p = 500$ predictors, a random forest with 50 trees requires storing only a $1000 \times 50$ matrix containing the leaf nodes for each observation in each tree, together with a vector of responses $y = (y_1, \dots, y_n)$.

\medskip
\noindent
\textbf{Key advantages.} Both methods retain the structural information captured by shallow forests while producing continuous, task-adapted weights. They add only a lightweight post-processing stage after tree traversal, and hyperparameter tuning over $h$ is computationally inexpensive, as forest structure and leaf assignments are precomputed once and reused across all evaluations. Importantly, \DiNo{} and \RanBu{} can be applied \emph{post hoc} to any pre-trained depth-limited forest without altering its training procedure, making them portable, easily integrable into existing RF-based workflows, and well-suited for latency-sensitive production environments.

\subsection{Quantile regression}
\label{ssec:qr}

Both \DiNo{} and \RanBu{} extend naturally to conditional quantile estimation \citep{koenker1978regression, Izbicki2025} by replacing the weighted mean with a weighted empirical cumulative distribution function (CDF). Let $w_i^{\bullet}(\mathbf{x})$ be the kernel-like weights defined in Section~\ref{ssec:kernelschemes} for method $\bullet \in \{\mathrm{DiNo}, \mathrm{RanBu}\}$. The weighted CDF is
\[
\hat F^{\bullet}(y\mid\mathbf{x}) =
\sum_{i=1}^n w_i^{\bullet}(\mathbf{x})\,\mathbb{I}\{Y_i\le y\},
\]
and the $\alpha$-quantile is obtained as
\[
\hat Q_{\alpha}^{\bullet}(\mathbf{x}) =
\inf\bigl\{y:\hat F^{\bullet}(y\mid\mathbf{x})\ge\alpha\bigr\}.
\]

This formulation assigns non-zero, smoothly decaying weights to all training points, in contrast to the discrete, leaf-restricted aggregation of Quantile Regression Forests~\citep{Meinshausen06quantileregression}. The result is a smoother conditional quantile function, which can improve stability in finite samples—especially for extreme quantiles—while retaining the computational efficiency of the shallow-forest design. Since leaf assignments are precomputed, quantile estimation becomes a post-processing step requiring no retraining, making it a straightforward extension of the mean estimators in Section~\ref{ssec:kernelschemes}.

\section{Simulation Results}
\label{sec:experiments}

\subsection{Simulation Scenarios}\label{sec:sim_scenarios}

We assess the predictive accuracy and computational efficiency of \DiNo{} and \RanBu{} in three scenarios—\textit{Rectangular Regions}, \textit{Friedman~1} \citep{Friedman91}, and \textit{Linear}. For each scenario, we run all combinations of three training sample sizes ($n\in\{500,\,1\,000,\,10\,000\}$) and three levels of irrelevant predictors ($R\in\{0,50,100\}$), isolating the effects of data volume and noise on model performance.

\subsubsection*{Rectangular Regions}
This scenario is designed to create a signal that is constant within rectangular regions of the covariate space, matching the type of axis–parallel partitions used by decision trees and random forests. The response is:
\begin{equation*}
\label{eq:hyperrectangle}
Y_i = -2.5\,X_{1i}(Z_{1i}) - 2\,X_{2i}(Z_{2i}) + 3\,X_{3i}(Z_{3i}) - 4\,X_{4i}(Z_{4i}) + \varepsilon_i,
\end{equation*}
where $Z_{ji}\sim\operatorname{Bernoulli}(0.5)$. Each covariate $X_{ji}$ is drawn uniformly from $(0,5)$ if $Z_{ji}=0$, and from $(5,10)$ if $Z_{ji}=1$. This construction produces inputs that fall into one of two disjoint intervals along each axis, so that the joint space is partitioned into axis–aligned rectangles. The error term is $\varepsilon_i\sim\mathcal N(0,1)$. In addition, $R$ independent noise variables are drawn from $\mathrm U(0,10)$ but have no effect on $Y$.

\subsubsection*{Friedman~1}
The nonlinear and interaction‐rich \textit{Friedman~1} function \citep{Friedman91} is:
\[
Y_i = 10\sin(\pi X_{1i}X_{2i}) + 20(X_{3i}-0.5)^2 + 10X_{4i} + 5X_{5i} + \varepsilon_i,
\]
with $X_{ji}\sim\mathrm U(0,1)$ for $j\le 5$ and $\varepsilon_i\sim\mathcal N(0,1)$. We extend the original formulation by adding $R$ independent noise variables $\sim\mathrm U(0,1)$.

\subsubsection*{Linear}
A correctly specified additive model:
\[
Y_i = 1X_{1i} + 3X_{2i} - 1.5X_{3i} - 2X_{4i} + 0.7X_{5i} + \varepsilon_i,
\]
where all predictors are drawn i.i.d.\ from $\mathcal N(0,1)$ and $\varepsilon_i\sim\mathcal N(0,1)$. As in other scenarios, $R$ additional Gaussian noise variables are included but unrelated to $Y$.

\subsection{Evaluation Metrics}

For conditional mean estimation, predictive accuracy is assessed using the mean squared error (MSE) on an independent test set of $n=1{,}000$ observations; 
\[
L(y,\hat{ y})=\frac{1}{n}\sum_{i=1}^{n}(y_i - \hat{y}_i)^2,
\]
where $y_i$ denotes the observed value and $\hat{y}_i$ the predicted value for the $i$-th observation. Thus, smaller values correspond to better predictive accuracy. Computational efficiency is measured as the wall‐clock time required to train the model and produce a single out‐of‐sample prediction, using the same hardware and software environment for all methods. Each experiment is replicated 50 times, and we report average results.

For conditional quantile estimation, we use the \emph{pinball loss} (\citet{koenker1978regression}):
\[
L_{\alpha}(y,\hat q_{\alpha})=\begin{cases}
 \alpha\,(y-\hat q_{\alpha}), & y\ge\hat q_{\alpha},\\
 (1-\alpha)(\hat q_{\alpha}-y), & y<\hat q_{\alpha},
 \end{cases}
\]
computed on the same test set and averaged over 99 quantiles ($\alpha=0.01,\dots,0.99$). Lower values again indicate better predictive accuracy.

\subsection{Methods and Hyper‐parameters}
We benchmark our two proposed methods, \DiNo{} and \RanBu{}, against four tree-based baselines using \textsf{R} implementations, keeping package defaults except where noted.

\begin{itemize}
    \item \textbf{Boosting}: this method builds an additive model in a stagewise manner, where at each iteration a new weak learner is fitted to the negative gradient of the loss function (the pseudo-residuals) of the current ensemble. This gradient-based update, combined with a shrinkage parameter, gradually refines the predictor. The implementation used was the \texttt{gbm} v2.2.2 package \citep{gbm2024}, with depth-1 trees (stumps), 100 trees, and shrinkage 0.1 (package defaults).  

    \item \textbf{GRF}: this method extends random forests by constructing adaptive local weights that solve target-specific moment equations, enabling estimation beyond the conditional mean (e.g., quantiles, treatment effects). A key feature is \emph{honest splitting}, where one subsample is used to build the partition and another to estimate local moments, which ensures asymptotic consistency and valid inference. The implementation used was the \texttt{grf} v2.4.0 package \citep{grf}, with 2,000 trees, honest splitting enabled, and no automatic tuning (\texttt{tune.parameters="none"}).

  \item \textbf{Random Forest (R.F.)}: \texttt{ranger} \citep{ranger} defaults (500 trees; no explicit \texttt{max.depth}); effective depth is bounded by stopping criteria such as \texttt{min.node.size} and \texttt{mtry}.
  \item \textbf{Reduced R.F.}: \texttt{ranger} with 50 trees and maximum depth 5 (chosen from preliminary runs and computational convenience), but without the weighting stage.
\end{itemize}

The proposed \DiNo{} and \RanBu{} estimators are both built on top of the Reduced R.F., adding a kernel-like weighting stage: \DiNo{} uses the MRCA distance, while \RanBu{} uses Breiman’s proximity, both with Gaussian bandwidth $h=0.2$.

All methods use the same train/test splits; keeping defaults reflects typical out-of-the-box usage and avoids biasing results via extra hyper-parameter tuning. 
Table~\ref{tab:definition:methods} summarizes the main hyper-parameters for each method.

\begin{table}[htbp]
\centering
\setlength{\tabcolsep}{8pt}
\begin{tabular}{lcccccc}
\toprule
\textbf{Argument} & \textbf{Boosting} & \textbf{\DiNo} & \textbf{GRF} & \textbf{\RanBu} & \textbf{R.F.} & \textbf{Reduced R.F.} \\
\midrule
Number of trees   & 100 & 50 & \(2\,000\) & 50 & 500 & 50 \\
Max depth         & 1   & 5  & \textemdash & 5  & unlimited$^{\dagger}$ & 5 \\
Shrinkage         & 0.1 & \textemdash & \textemdash & \textemdash & \textemdash & \textemdash \\
Bandwidth $h$     & \textemdash & 0.2 & \textemdash & 0.2 & \textemdash & \textemdash \\
\bottomrule
\end{tabular}
\caption{Hyper-parameters used for each method. \textemdash\ indicates not applicable. $^{\dagger}$In \texttt{ranger}, no explicit \texttt{max.depth} is set; effective depth is bounded by stopping criteria such as \texttt{min.node.size} and \texttt{mtry}.}
\label{tab:definition:methods}
\end{table}

\subsection{Conditional Expectation Results}
\label{sec:results_mean}

Tables~\ref{tab:eqm:simulation} and~\ref{tab:eqm:simulation:time} in the Appendix report mean squared error (MSE) and runtime ratios across all simulated scenarios. Ratios below~1 favour the method in the row, whereas ratios above~1 favour the method in the column. To aid interpretation, pairwise outcomes are further classified into four categories—(i) \emph{lower loss / lower time}, (ii) \emph{lower loss / higher time}, (iii) \emph{higher loss / lower time}, and (iv) \emph{higher loss / higher time}—and summarised in Table~\ref{tab:resumo:simul:tempo}.

Across scenarios, four consistent patterns emerge. In high-noise regimes, \RanBu{} achieves simultaneous gains in accuracy and efficiency, while \DiNo{} exchanges a moderate runtime penalty for stable error reduction. When noise is absent, differences shrink and R.F.\ remains competitive for small $n$, yet the proposed methods continue to deliver measurable error reductions as sample size grows. In the linear setting, \DiNo{} is preferable at moderate $n$, whereas \RanBu{} dominates once irrelevant variables are introduced. Most importantly, both methods substantially improve upon the Reduced R.F., demonstrating that the kernel-like weighting stage enhances the representational capacity of the shallow forest rather than merely smoothing its predictions.

\paragraph{Key findings.}
\begin{itemize}
    \item \textbf{High‐noise settings ($R=50$ or $100$):} \RanBu{} reduces MSE by 40--70\% relative to R.F.\ and GRF while cutting runtime by up to 10–50 times. \DiNo{} matches the accuracy of R.F.\ with a runtime overhead of 20--40\%.
    \item \textbf{Low‐noise settings ($R=0$):} differences narrow; R.F.\ remains competitive for small $n$. In the rectangular design, however, both \RanBu{} and \DiNo{} achieve 5--10\% error reductions for $n \geq 5{,}000$.
    \item \textbf{Linear setting:} \DiNo{} yields the lowest errors at moderate sample sizes ($n=1{,}000$--$5{,}000$), with MSE ratios around 0.7 compared to R.F.; as noise grows, \RanBu{} dominates with both lower error (up to 30\%) and faster runtime.
    \item \textbf{Against Reduced R.F.:} both methods improve accuracy dramatically (often by factors of 2--3 in high-noise scenarios) while incurring only modest runtime penalties ($\sim$20--50\%). This confirms that the weighting stage is not just smoothing, but structurally strengthens the reduced forest baseline into a competitive predictor.
\end{itemize}

\begin{table}[h]
\centering
\begin{tabular}{lllcccc}
\hline
\cellcolor[HTML]{EFEFEF}\textbf{Method} & \cellcolor[HTML]{EFEFEF}\textbf{Compared to} & \cellcolor[HTML]{EFEFEF}\textbf{Noise} & \cellcolor[HTML]{67FD9A}\textbf{\begin{tabular}[c]{@{}c@{}}Lower Loss \\and\\ Lower time\end{tabular}} & \cellcolor[HTML]{FFFE65}\textbf{\begin{tabular}[c]{@{}c@{}} Lower loss \\and\\ Greater time\end{tabular}} & \cellcolor[HTML]{FFFE65}\textbf{\begin{tabular}[c]{@{}c@{}}Greater Loss \\and\\ Lower time\end{tabular}} & \cellcolor[HTML]{FD6864}\textbf{\begin{tabular}[c]{@{}c@{}}Greater Loss \\and\\ Greater time\end{tabular}} \\ \hline
\multirow{9}{*}{\shortstack{\DiNo}}  & \multirow{3}{*}{R.F.}        &   0 & 2 (16.7\%)   &  2 (16.7\%)   &  4 (33.3\%) & 4 (33.3\%) \\                     
                                    &                               &  50 & 5 (41.7\%)   &  3 (25.0\%)   &  4 (33.3\%) & -          \\                     
                                    &                               & 100 & 6 (50.0\%)   &  3 (25.0\%)   &  3 (25.0\%) & -          \\ \cline{2-7}                  
                                    & \multirow{3}{*}{Reduced R.F.} &   0 & -            &  12 (100.0\%) &  -          & -          \\                     
                                    &                               &  50 & -            &  12 (100.0\%) &  -          & -          \\                     
                                    &                               & 100 & -            &  12 (100.0\%) &  -          & -          \\ \cline{2-7}                     
                                    & \multirow{3}{*}{GRF}          &   0 & 3 (25.0\%)   &  3 (25.0\%)   &  6 (50.0\%) & -          \\                     
                                    &                               &  50 & 5 (41.7\%)   &  -            &  7 (58.3\%) & -          \\                     
                                    &                               & 100 & 6 (50.0\%)   &  -            &  6 (50.0\%) & -          \\ \hline

\multirow{9}{*}{\shortstack{\RanBu}} & \multirow{3}{*}{R.F.}        &   0 & 3 (25.0\%)   &  -            &  9 (75.0\%) & -          \\                     
                                    &                               &  50 & 12 (100.0\%) &  -            &  -          & -          \\                     
                                    &                               & 100 & 12 (100.0\%) &  -            &  -          & -          \\   \cline{2-7}                
                                    & \multirow{3}{*}{Reduced R.F.} &   0 & -            &  11 (91.7\%)  &  -          & 1 ( 8.3\%) \\                     
                                    &                               &  50 & -            &  12 (100.0\%) &  -          & -          \\                     
                                    &                               & 100 & -            &  12 (100.0\%) &  -          & -          \\ \cline{2-7}                   
                                    & \multirow{3}{*}{GRF}          &   0 & 8 (66.7\%)   &  -            &  4 (33.3\%) & -          \\                     
                                    &                               &  50 & 8 (66.7\%)   &  -            &  4 (33.3\%) & -          \\                     
                                    &                               & 100 & 9 (75.0\%)   &  -            &  3 (25.0\%) & -          \\ \hline                  

\end{tabular}
\caption{Summary of simulation outcomes: number (and percentage) of cases in each performance category when comparing \DiNo{} and \RanBu{} to baseline methods. Green cells indicate simultaneous improvements in accuracy and speed, while red marks degradation in both. Results show that \DiNo{} consistently improves over the Reduced R.F.\ at the cost of longer runtimes, whereas \RanBu{} delivers frequent dual improvements, particularly against GRF, confirming the practical advantage of the kernel weighting scheme. Importantly, both methods tend to perform especially well in high-noise settings, highlighting their robustness to irrelevant features.}
    \label{tab:resumo:simul:tempo}
\end{table}

\subsection{Runtime Scaling}
\begin{figure}[h]
    \centering
    \includegraphics[width=1\linewidth]{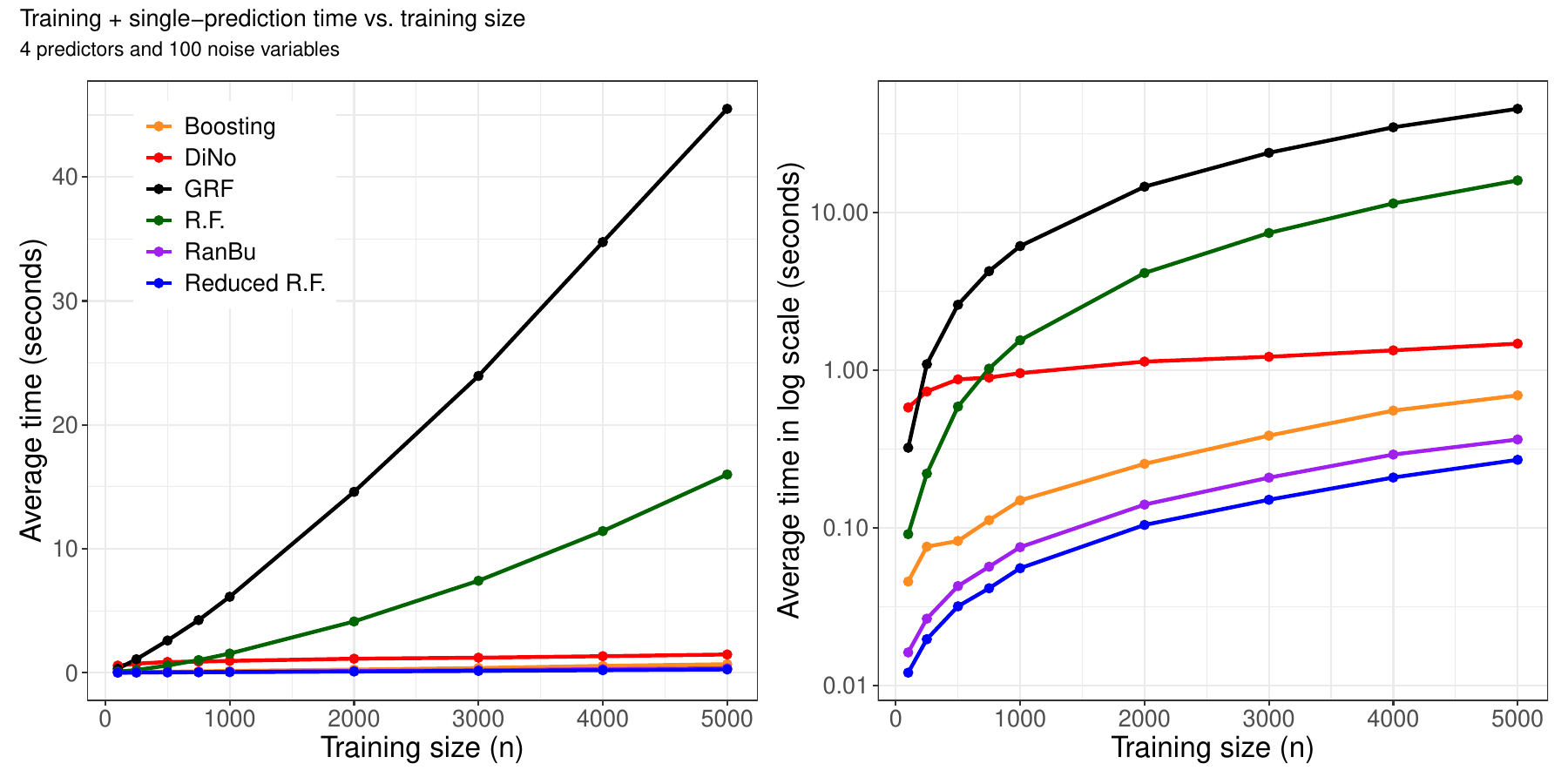}    
    \caption{
Average time (in seconds) required to train each model and generate a prediction for a single test observation, as a function of the training set size. The experiment is based on the \textit{friedman} simulation setting with 4 informative predictors and 100 noise variables. \RanBu{} and \DiNo{} remain efficient even as sample size increases, while full-depth Random Forests (\textsc{R.F.}) and GRF exhibit substantially higher computational costs. The right-hand panel uses a logarithmic scale on the $y$-axis to better visualize differences across methods. Results are averaged over 50 replications.
} 
    \label{fig:tempo_processamento}
\end{figure}

Table~\ref{tab:resumo:simul:tempo} summarises win/loss categories, and Figure~\ref{fig:tempo_processamento} shows runtime growth with $n$. GRF scales poorest, followed by R.F. Reduced R.F.\ is fastest overall, but \RanBu{} is close, and remains much faster than R.F./GRF for large $n$. \DiNo{} starts slower on small $n$ but overtakes R.F.\ and GRF beyond $n \approx 750$. The Boosting timings reflect a fixed hyperparameter setting; cross‐validated tuning would increase its cost substantially.

\subsection{Quantile Regression Results}
\label{sec:results_quantile}

Tables~\ref{tab:pinball:simulation} and~\ref{tab:pinball:simulation:time} report pinball loss and runtime ratios, evaluated as the average pinball loss across quantiles $0.01, 0.02, \ldots, 0.99$. A condensed overview appears in Table~\ref{tab:resumo:simul:tempo:quantile}. The relative ordering of methods is highly consistent with the conditional mean case: settings where \RanBu{} and \DiNo{} improved over the baselines in mean regression also show corresponding gains in quantile regression. This suggests that the tree‐based distances underlying the proposed estimators provide a stable representation that extends naturally from conditional expectation to conditional distribution estimation.

In high‐noise scenarios ($R=50$ or $100$), \RanBu{} consistently outperforms R.F.\ and GRF, achieving lower pinball loss while often reducing runtime as well. \DiNo{} also improves accuracy in these regimes, though its gains are typically accompanied by higher computational cost. Under low noise ($R=0$), differences across methods are smaller: \RanBu{} retains moderate advantages over GRF and Reduced R.F., while \DiNo{} shows mixed behavior, with some improvements in accuracy offset by slower execution.

In the \textit{Linear} scenario, \DiNo{} achieves competitive pinball loss reductions, particularly at moderate sample sizes, but again tends to incur additional runtime relative to the baselines. Across all designs, both \DiNo{} and \RanBu{} consistently improve upon their underlying Reduced R.F.\ baseline, confirming that the similarity‐weighted estimators enhance the predictive capacity of reduced forests in conditional distribution estimation.

Runtime scaling trends are broadly analogous to those observed for mean regression. GRF exhibits the steepest growth with $n$, while Reduced R.F.\ remains the fastest baseline. \RanBu{} lies close behind, combining computational efficiency with robust accuracy. By contrast, \DiNo{} is slower in small samples but eventually surpasses R.F.\ and GRF runtimes for $n \gtrsim 750$, maintaining favourable scaling for larger data sets. As in the conditional expectation case, Boosting results are reported for a fixed hyperparameter configuration; cross‐validated tuning would substantially increase its computational burden.

\begin{table}[h]
\centering
\begin{tabular}{lllcccc}
\hline
\cellcolor[HTML]{EFEFEF}\textbf{Method} & \cellcolor[HTML]{EFEFEF}\textbf{Compared to} & \cellcolor[HTML]{EFEFEF}\textbf{Noise} & \cellcolor[HTML]{67FD9A}\textbf{\begin{tabular}[c]{@{}c@{}}Lower Loss \\and\\ Lower time\end{tabular}} & \cellcolor[HTML]{FFFE65}\textbf{\begin{tabular}[c]{@{}c@{}} Lower loss \\and\\ Greater time\end{tabular}} & \cellcolor[HTML]{FFFE65}\textbf{\begin{tabular}[c]{@{}c@{}}Greater Loss \\and\\ Lower time\end{tabular}} & \cellcolor[HTML]{FD6864}\textbf{\begin{tabular}[c]{@{}c@{}}Greater Loss \\and\\ Greater time\end{tabular}} \\ \hline

\multirow{9}{*}{\shortstack{\DiNo}}  & \multirow{3}{*}{R.F.}         &   0 & -              &   -             &  6 (50.0\%)   &  6 (50.0\%)   \\               
                                     &                               &  50 & 1 ( 8.3\%)     &   1 ( 8.3\%)    &  8 (66.7\%)   &  2 (16.7\%)   \\               
                                     &                               & 100 & 1 ( 8.3\%)     &   1 ( 8.3\%)    &  8 (66.7\%)   &  2 (16.7\%)   \\    \cline{2-7}            
                                     & \multirow{3}{*}{Reduced R.F.} &   0 & -              &   -             &  -            &  12 (100.0\%) \\                     
                                     &                               &  50 & -              &   8 (66.7\%)    &  -            &  4 (33.3\%)   \\               
                                     &                               & 100 & -              &   9 (75.0\%)    &  -            &  3 (25.0\%)   \\    \cline{2-7}                  
                                     & \multirow{3}{*}{GRF}          &   0 & 1 ( 8.3\%)     &   2 (16.7\%)    &  8 (66.7\%)   &  1 ( 8.3\%)   \\               
                                     &                               &  50 & 8 (66.7\%)     &   -             &  4 (33.3\%)   &  -            \\               
                                     &                               & 100 & 8 (66.7\%)     &   -             &  4 (33.3\%)   &  -            \\   \hline

\multirow{9}{*}{\shortstack{\RanBu}} & \multirow{3}{*}{R.F.}          &   0 & 3 (25.0\%)     &   -             &  9 (75.0\%)   &  -            \\               
                                     &                                &  50 & 12 (100.0\%)   &   -             &  -            &  -            \\               
                                     &                                & 100 & 12 (100.0\%)   &   -             &  -            &  -            \\  \cline{2-7}                    
                                     & \multirow{3}{*}{Reduced R.F.}  &   0 & -              &   10 (83.3\%)   &  -            &  2 (16.7\%)   \\               
                                     &                                &  50 & -              &   12 (100.0\%)  &  -            &  -            \\               
                                     &                                & 100 & -              &   12 (100.0\%)  &  -            &  -            \\   \cline{2-7}                   
                                     & \multirow{3}{*}{GRF}           &   0 & 12 (100.0\%)   &   -             &  -            &  -            \\               
                                     &                                &  50 & 12 (100.0\%)   &   -             &  -            &  -            \\               
                                     &                                & 100 & 12 (100.0\%)   &   -             &  -            &  -            \\   \hline

\end{tabular}
\caption{Summary of quantile regression simulation outcomes: number (and percentage) of cases in each performance category when comparing \DiNo{} and \RanBu{} to baseline methods. Green cells indicate simultaneous improvements in accuracy and speed, while red marks degradation in both. Results show that \DiNo{} underperforms the Reduced R.F.\ in noise–free settings, but gains a consistent advantage as irrelevant predictors are introduced, highlighting robustness to noisy covariates. In contrast, \RanBu{} frequently secures dual improvements, particularly over GRF, and maintains stable performance across all noise levels.}

    \label{tab:resumo:simul:tempo:quantile}
\end{table}

\section{Real-World Applications}
\label{sec:applications}

We now assess the predictive accuracy and computational efficiency of \DiNo{} and \RanBu{} on the real‐world datasets listed in Table~\ref{tab_descricao_dados}, following the same evaluation protocol described in the last section. Each method is compared against four baselines: Boosting, Generalized Random Forests (GRF), a Reduced Random Forest (Reduced R.F.), and a full Random Forest (R.F.). 

Predictive accuracy is measured via \emph{relative loss} (values $<1$ indicate improvement over the baseline), and efficiency is quantified as \emph{relative runtime} (smaller is better). A summary of win/loss categories appears in Table~\ref{tab:resumo:simul:tempo:app}. Beyond simple win counts, we also report average ratios across datasets to better capture the magnitude of improvements (\textbf{Tables} 
\ref{tab:eqm:application_ratio}, \ref{tab:time:application_ratio}, \ref{tab:pinball:application_ratio}, and \ref{tab:time:application_ratio:quantile}).

\begin{table}[h]
\centering
\begin{tabular}{lllcccc}
\hline
\cellcolor[HTML]{EFEFEF}\textbf{Context} & \cellcolor[HTML]{EFEFEF}\textbf{Method} & \cellcolor[HTML]{EFEFEF}\textbf{Compared to} & \cellcolor[HTML]{67FD9A}\textbf{\begin{tabular}[c]{@{}c@{}}Lower Loss \\and\\ Lower time\end{tabular}} & \cellcolor[HTML]{FFFE65}\textbf{\begin{tabular}[c]{@{}c@{}} Lower loss \\and\\ Greater time\end{tabular}} & \cellcolor[HTML]{FFFE65}\textbf{\begin{tabular}[c]{@{}c@{}}Greater Loss \\and\\ Lower time\end{tabular}} & \cellcolor[HTML]{FD6864}\textbf{\begin{tabular}[c]{@{}c@{}}Greater Loss \\and\\ Greater time\end{tabular}} \\ \hline
\multirow{8}{*}{\shortstack{Conditional \\ Expectation}} &   \multirow{4}{*}{\DiNo}              & Boosting     & -           & 13 (56.5\%)  & -           & 10 (43.5\%) \\
                                                         &                                       & GRF          & 7 (30.4\%)  & 4 (17.4\%)   & 9 (39.1\%)  & 3 (13.0\%)  \\
                                                         &                                       & Reduced R.F. & -           & 16 (69.6\%)  & -           & 7 (30.4\%)  \\
                                                         &                                       & R.F.         & -           & 2 (8.7\%)    & 3 (13.0\%)  & 18 (78.3\%) \\ \cline{2-7} 
                                                         &  \multirow{4}{*}{\shortstack{\RanBu}} & Boosting     & 8 (30.8\%)  & 7 (26.9\%)   & 6 (23.1\%)  & 5 (19.2\%)  \\
                                                         &                                       & GRF          & 15 (57.7\%) & -            & 10 (38.5\%) & 1 (3.8\%)   \\
                                                         &                                       & Reduced R.F. & 1 (3.8\%)   & 17 (65.4\%)  & -           & 8 (30.8\%)  \\
                                                         &                                       & R.F          & 4 (15.4\%)  & -            & 22 (84.6\%) & -           \\ \hline

\multirow{6}{*}{\shortstack{Quantile}}  & \multirow{3}{*}{\DiNo}               & GRF          & 11 (50.0\%) & 3 (13.6\%)  & 8 (36.4\%)  & -           \\
                                        &                                      & Reduced R.F. & -           & 20 (87.0\%) & -           & 3 (13.0\%)  \\
                                        &                                      & R.F.         & -           & 1 (4.3\%)   & 5 (21.7\%)  & 17 (73.9\%) \\ \cline{2-7}
                                        & \multirow{3}{*}{\shortstack{\RanBu}} & GRF          & 12 (48.0\%) & -           & 12 (48.0\%) & 1 (4.0\%)   \\
                                        &                                      & Reduced R.F. & 1 (3.8\%)   & 17 (65.4\%) & -           & 8 (30.8\%)  \\
                                        &                                      & R.F          & 1 (3.8\%)   & -           & 25 (96.2\%) & -           \\ \hline
\end{tabular}
\caption{Summary of results on the real–world data sets listed in Table~\ref{tab_descricao_dados}. Green cells indicate simultaneous gains in accuracy and speed, red marks degradation in both. For conditional expectation, \RanBu{} consistently outperforms GRF in more than half of the cases and also shows advantages over Boosting and Reduced R.F., whereas \DiNo{} often trades accuracy for speed, and performs poorly against the full R.F. In the quantile setting, \DiNo{} shows mixed results—competing well with GRF but underperforming relative to R.F.—while \RanBu{} achieves the strongest improvements overall, particularly against GRF, confirming its robustness across diverse data sets.}

    \label{tab:resumo:simul:tempo:app}
\end{table}

\subsection{Conditional Expectation}
\RanBu{} shows a consistent advantage over GRF, achieving both lower loss and faster execution in 57.7\% of datasets, with an average relative loss of $0.94$ and runtime ratio of $0.87$. Relative to Boosting, it improves predictive accuracy in 57.7\% of cases (average loss ratio $0.97$), with simultaneous gains in both metrics in 30.8\%. Against Reduced R.F., \RanBu{} improves both predictive accuracy and computational efficiency in 65.4\% of tasks (average loss ratio $0.92$), demonstrating that the kernel weighting scheme strengthens the reduced forest baseline rather than merely pruning trees. Comparisons with full R.F.\ reveal fewer joint wins (15.4\%), but \RanBu{} remains faster in the majority of cases, with average runtimes $25\%$ lower.

\DiNo{} tends to incur higher runtimes than Boosting (average runtime ratio $1.18$), but still improves accuracy in 56.0\% of datasets (average loss ratio $0.96$). When compared with GRF, it delivers dual gains in 30.4\% of cases and higher accuracy in nearly half. Against Reduced R.F., \DiNo{} achieves substantial improvements (69.6\% of tasks, average loss ratio $0.91$). In head-to-heads with full R.F., wins are rare (8.7\%), and most losses occur when both metrics worsen.

Overall, these results underscore that the kernel weighting scheme is not a trivial extension of forest-based proximities. By transforming tree-based distances into continuous similarity weights, both \DiNo{} and \RanBu{} extract structural information that simple forest pruning cannot, providing real predictive value beyond reduced ensembles.

\subsection{Conditional Quantiles}
For quantile regression, \RanBu{} outperforms GRF in both loss and time in 48.0\% of datasets, with average loss ratio $0.95$. Reduced R.F.\ remains competitive in speed, but \RanBu{} still achieves better predictive accuracy in 69.8\% of cases, even when runtime is higher. Against full R.F., accuracy gains are rare, but the method still offers consistent runtime savings.

\DiNo{} matches or surpasses GRF in half the datasets for both metrics, with average loss ratio $0.96$, and beats Reduced R.F.\ in 87.0\% of cases (average loss ratio $0.90$), albeit with increased runtime. In head-to-head comparisons with full R.F., wins are infrequent, particularly due to runtime penalties.

Overall, these results confirm that the kernel weighting scheme is not a cosmetic addition but a substantive enhancement. By converting tree-based distances into continuous similarity weights, both \DiNo{} and \RanBu{} extract richer structural information than naive forest reductions, yielding systematic improvements in predictive performance across quantiles.

\subsection{Loss Across Quantiles}
\begin{figure}[h]
    \centering
    \includegraphics[width=1\linewidth]{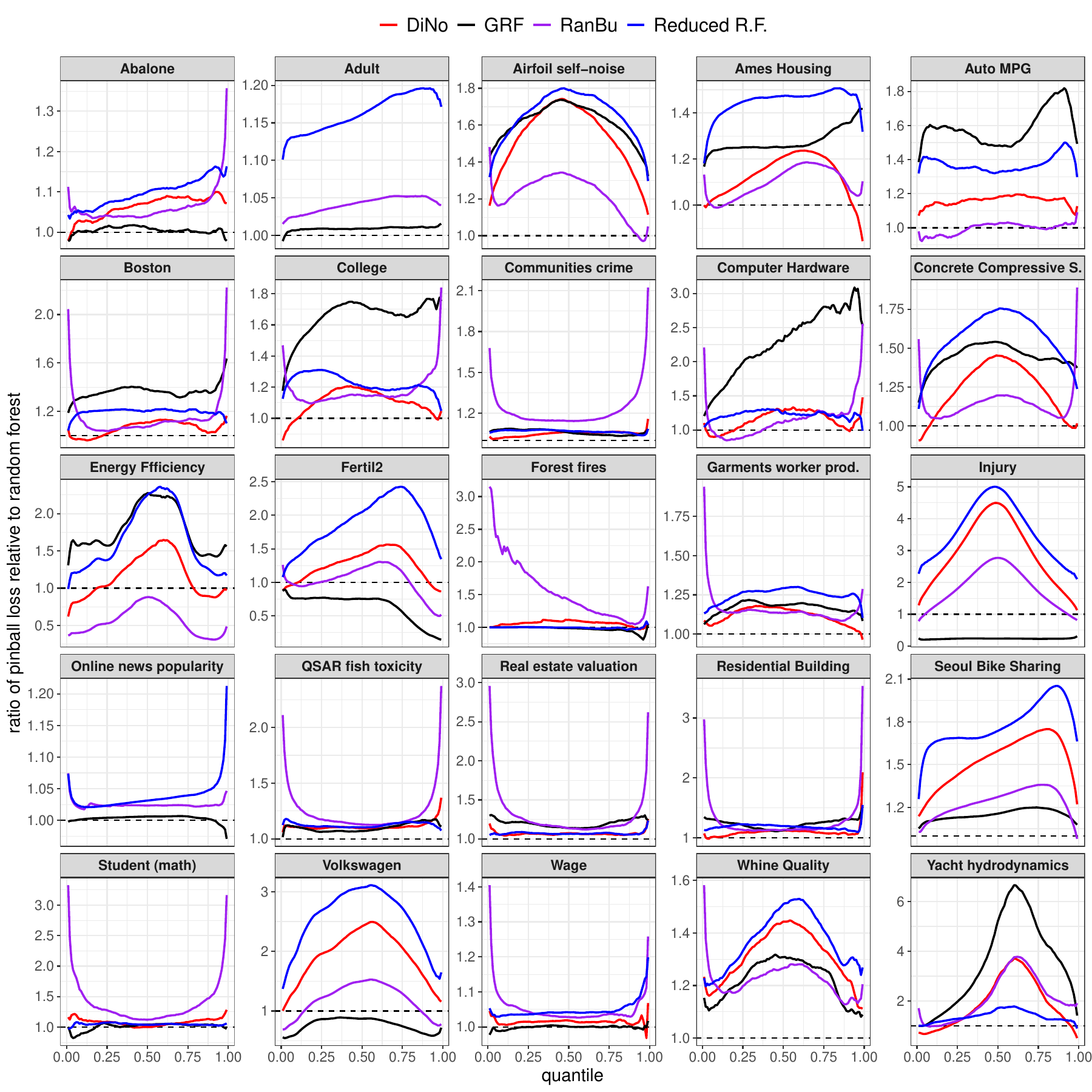}    
   \caption{Ratio of pinball loss to Random Forest across quantiles for \DiNo{}, \RanBu{}, GRF, and Reduced Random Forest. 
Bandwidth for \DiNo{} and \RanBu{} fixed at $h=0.20$. Across most datasets, \DiNo{} (red) and \RanBu{} (blue) achieve ratios close to or below one, indicating comparable or superior accuracy to the full R.F.\ baseline. GRF (black) and Reduced R.F.\ (purple) often yield higher losses, especially at intermediate quantiles. The robustness of \DiNo{} and \RanBu{} across diverse distributions highlights their suitability for quantile regression tasks.}

    \label{fig:pinb:application:quantiles}
\end{figure}

Figure~\ref{fig:pinb:application:quantiles} shows pinball loss ratios relative to R.F.\ across quantiles. Both \DiNo{} and \RanBu{}, with $h=0.20$, frequently achieve ratios $\leq 1$, especially in central quantiles. Reduced R.F.\ generally fares worse, while GRF performance varies widely—sometimes excelling (e.g., \emph{Abalone}, \emph{Fertil}, \emph{Volkswagen}, and \emph{QSAR fish toxicity}) but often degrading, particularly in datasets like \emph{Auto MPG} and \emph{Yacht hydrodynamics}. Performance tends to be more variable in the extreme quantiles, with no single method dominating systematically across all datasets.

\subsection{Sensitivity to Bandwidth}
\begin{figure}[h]
    \centering
    \includegraphics[width=1\linewidth]{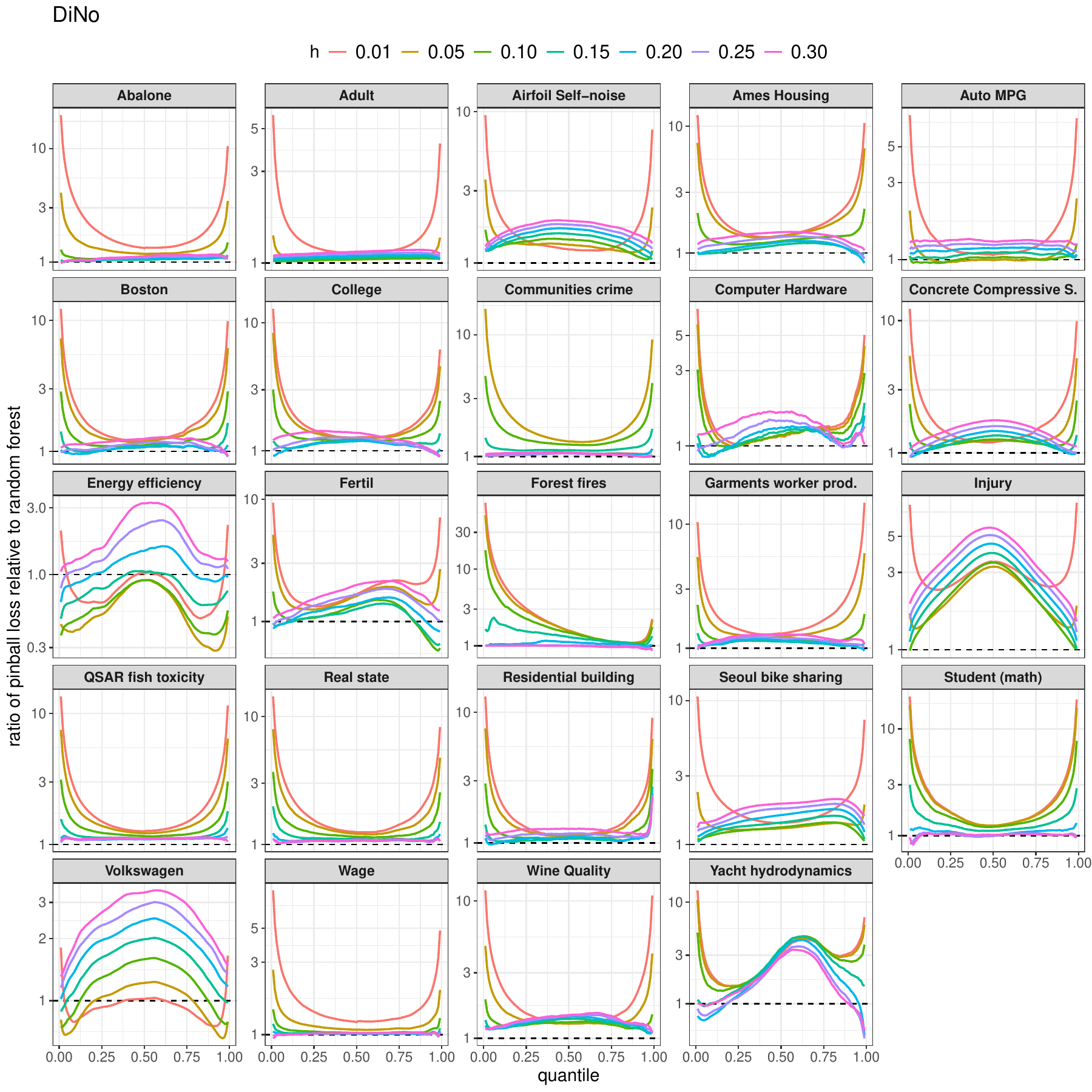}    
\caption{Ratio of pinball loss of \DiNo{} to Random Forest across quantiles for different bandwidth values $h$. 
Performance is sensitive to bandwidth choice: very small $h$ inflates loss at the distribution tails, while very large $h$ tends to degrade accuracy at central quantiles. Intermediate values (e.g., $h=0.15$–$0.20$) generally yield the most stable and competitive performance across datasets.}
    \label{fig:pinb:application:quantiles_dino}
\end{figure}

\begin{figure}[h]
    \centering
    \includegraphics[width=1\linewidth]{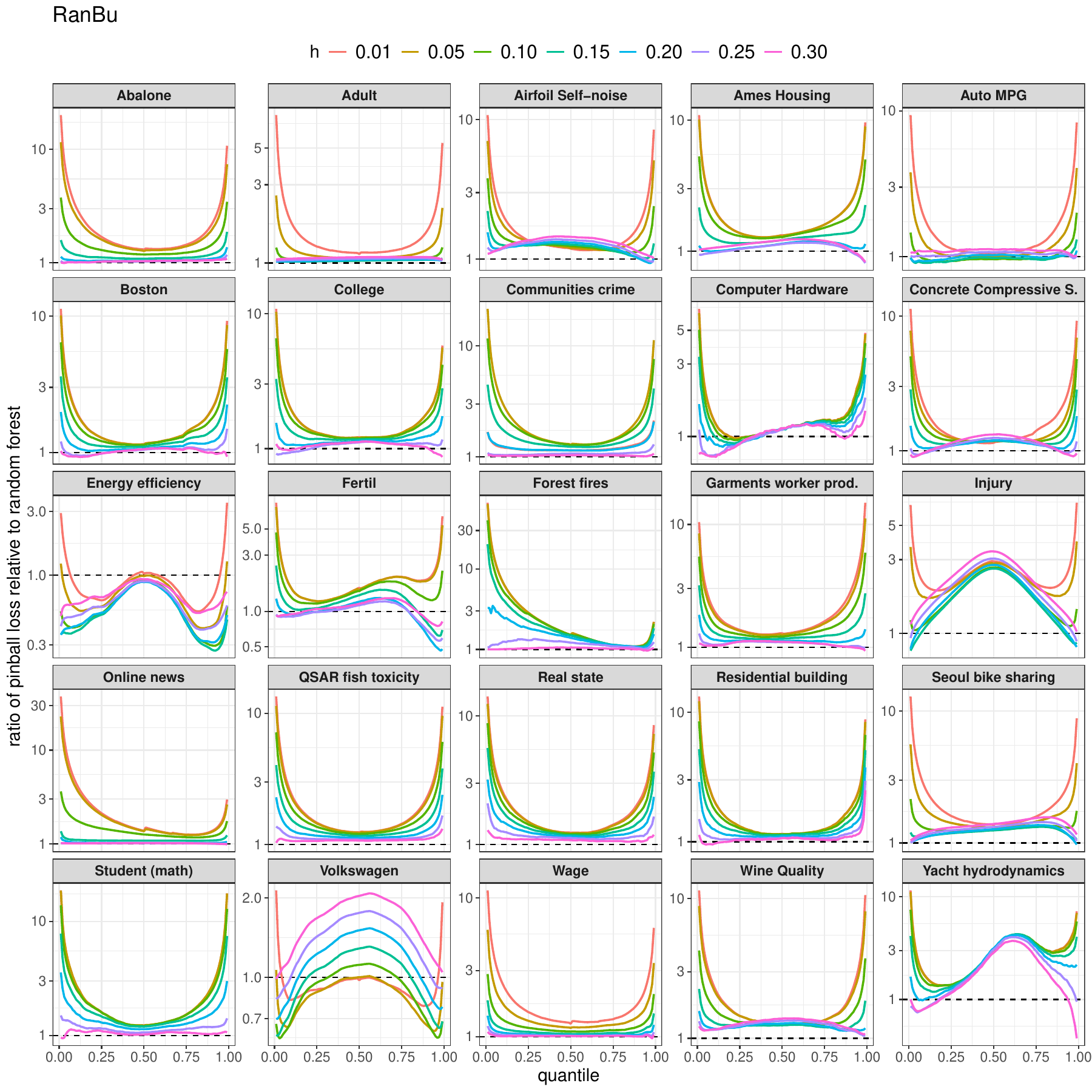}    
\caption{Ratio of pinball loss of \RanBu{} to Random Forest across quantiles for different bandwidth values $h$. 
Curves are generally closer to one, indicating greater robustness to bandwidth choice and more consistent performance across quantiles. 
Unlike \DiNo{}, \RanBu{} is less affected by extreme values of $h$, with stable behavior across datasets and competitive results even for non-optimal bandwidths.}

    \label{fig:pinb:application:quantiles_ranbu}
\end{figure}

Figures~\ref{fig:pinb:application:quantiles_dino}--\ref{fig:pinb:application:quantiles_ranbu} explore the effect of varying $h$. A ``U-shaped'' pattern emerges: very small $h$ values (0.01--0.05) overfit, inflating tail losses; large $h$ values (0.25--0.30) oversmooth and degrade central quantiles. 

\DiNo{} is more sensitive, with small $h$ sometimes causing severe tail degradation and large $h$ harming central quantiles (\emph{e.g.}, \emph{Garments worker productivity}). \RanBu{} maintains curves closer to~1 over a broader $h$ range, indicating greater robustness and more stable performance across datasets. Intermediate $h$ values (0.10--0.20) generally provide the best trade-off for both methods.

\subsection{Discussion}
In real-world tasks, \DiNo{} and \RanBu{} frequently improve predictive accuracy over GRF and Reduced R.F., with average loss reductions of roughly $5$--$10\%$. The consistent advantage over Reduced R.F.\ is particularly revealing: improvements cannot be explained by forest size alone, but stem from the kernel weighting scheme, which transforms tree-based distances into informative similarity weights. Overall, the results confirm the practical appeal of kernelized tree distances for both mean and quantile prediction, offering robust performance, predictable behavior, and competitive runtimes across diverse tasks.

\section{Final Remarks}
\label{sec:conclusion}

This paper introduced two new algorithms—\DiNo{} and \RanBu{}—that connect two established but often separate lines of research in supervised learning: tree–based ensembles and kernel–based similarity methods. Both approaches exploit the hierarchical structure of decision trees to define distances, which are then transformed into similarity weights through a kernel scheme. In doing so, they preserve the practical advantages of forests—minimal preprocessing, natural handling of mixed data types, and built–in feature selection—while adding a principled metric structure that can be directly exploited in prediction.

\subsection*{Key Findings}
\begin{itemize}
    \item Across 25 benchmark datasets (Table~\ref{tab_descricao_dados}), \RanBu{} outperformed GRF in predictive accuracy in more than half of the cases and was consistently faster, achieving its dual goal of accuracy and efficiency.
    \item Against Gradient Boosting, \RanBu{} improved accuracy in over half of the datasets and achieved joint gains in accuracy and runtime in nearly one–third.
    \item \DiNo{}, despite relying on heavily pruned trees, surpassed Reduced R.F.\ in roughly two–thirds of the tasks, confirming that the kernel weighting step provides predictive value beyond simple pruning.
    \item Both algorithms produced stable, well–scaled similarity scores, suitable for downstream tasks such as conformal prediction or anomaly detection without additional calibration.
    \item In simulation studies, both methods remained competitive even with large numbers of noise variables, a scenario that mirrors real–world practice where irrelevant features are abundant.
\end{itemize}

The methods are implemented in an open–source \textsf{R} package with syntax aligned to \texttt{ranger} and \texttt{grf}, lowering the barrier to adoption. Once a forest has been trained, its partition structure can be reused without retraining: distances are extracted from the fixed trees and converted into similarity weights that can be applied to new datasets. This design enables fast updating of predictions when either new sample points or alternative response variables \(Y\) become available, while avoiding the cost of growing additional trees. In practice, this allows practitioners to integrate predictive modeling with similarity–based analysis in a seamless way, reusing the same shallow forest backbone across tasks. Although this does not in itself solve dataset shift or domain adaptation, it provides a lightweight mechanism for post–hoc reuse of trained forests that can be valuable in dynamic or multi–label applications. The empirical results across diverse datasets indicate that these approaches offer a robust balance of accuracy and runtime, making them attractive for applied use.

Our experiments focused on structured, tabular data, where tree ensembles are most commonly used. Applications to high–dimensional unstructured domains such as images, text, or audio—where deep representation learning is dominant—remain unexplored. Furthermore, while results suggest strong empirical performance, additional work is needed to study statistical properties and to assess performance under more complex data–generating processes, such as clustered or longitudinal settings.

The proposed ideas can be naturally extended to other contexts. For classification, both estimators can be adapted by replacing the weighted mean with a weighted majority vote. Beyond prediction, the proposed distances are also of intrinsic interest: by formalising the geometry induced by tree ensembles, they provide a principled basis for similarity–driven tasks such as clustering or manifold exploration. Figure~\ref{fig:ames:distances} in the Appendix illustrates this point with a subset of the \emph{Ames Housing} data: applying multidimensional scaling (MDS) to the distance matrices reveals that \DiNo{} produces a well–spread embedding, while Breiman distances from a default random forest collapse most pairs to values near one, and the Reduced R.F.\ yields an intermediate structure. The corresponding histograms highlight this contrast, with \DiNo{} generating a broader distribution of pairwise distances, which suggests stronger discriminatory power. These properties indicate that our distances can serve as stand–alone tools for unsupervised analysis.

Tree ensembles have long been valued for their predictive accuracy and robustness. By integrating their induced geometry into a kernel weighting scheme, \DiNo{} and \RanBu{} turn forests into similarity–aware learners that are accurate, scalable, and easy to deploy. Their resilience to noise variables is particularly appealing in practice, where high–dimensional and heterogeneous data are the norm. We hope these contributions strengthen the connection between metric learning and ensemble methods, translating methodological advances into practical benefits across a wide range of applications.

\section*{Acknowledgements}

RI is grateful for the financial support of FAPESP (grant 2023/07068-1) and CNPq (grants 422705/2021-7 and 305065/2023-8).

\newpage
\bibliographystyle{abbrvnat}
\bibliography{references}  

\newpage
\appendix

\section*{Appendix}
\label{section:appendix}

\begin{table}[h]
\centering
\caption{Mean and standard error of MSE ratios comparing each method to \RanBu{} and \DiNo{}. Ratios $>1$ indicate improved performance of the proposed methods. Results show that both estimators consistently outperform baselines across scenarios, particularly in small training sizes and with many noise variables, confirming robustness to high-dimensional irrelevant features.}
\label{tab:eqm:simulation}
\footnotesize
\begin{tabular}{lllcccccc}
\toprule
\multirow{2}{*}{Scenario} & \multirow{2}{*}{Method} & \multirow{2}{*}{$n_{\text{train}}$} & \multicolumn{3}{c}{\DiNo} & \multicolumn{3}{c}{\RanBu} \\
\cmidrule(lr){4-6} \cmidrule(lr){7-9}
 &  & & Noise 0 & Noise 50 & Noise 100 & Noise 0 & Noise 50 & Noise 100 \\
\midrule

rectangular & R.F.         &  500  & \textbf{1.09 (0.01)} & \textbf{2.31 (0.03)} & \textbf{2.64 (0.05)} & 0.93 (0.00)          & \textbf{2.17 (0.02)} & \textbf{2.60 (0.03)} \\    
           &              & 1000  & \textbf{1.12 (0.00)} & \textbf{2.12 (0.02)} & \textbf{2.42 (0.03)} & \textbf{1.03 (0.00)} & \textbf{2.06 (0.02)} & \textbf{2.64 (0.02)} \\    
           &              & 5000  & \textbf{1.08 (0.00)} & \textbf{1.40 (0.01)} & \textbf{1.64 (0.02)} & \textbf{1.07 (0.00)} & \textbf{1.45 (0.01)} & \textbf{1.96 (0.01)} \\    
           &              & 10000 & \textbf{1.08 (0.00)} & \textbf{1.25 (0.01)} & \textbf{1.41 (0.02)} & \textbf{1.08 (0.00)} & \textbf{1.31 (0.01)} & \textbf{1.63 (0.01)} \\ \cline{2-9}     
           & Reduced R.F. &  500  & \textbf{1.11 (0.01)} & \textbf{2.95 (0.04)} & \textbf{3.27 (0.05)} & 0.96 (0.01)          & \textbf{2.77 (0.03)} & \textbf{3.22 (0.04)} \\    
           &              & 1000  & \textbf{1.10 (0.00)} & \textbf{3.23 (0.04)} & \textbf{3.52 (0.04)} & \textbf{1.01 (0.00)} & \textbf{3.14 (0.03)} & \textbf{3.85 (0.04)} \\    
           &              & 5000  & \textbf{1.06 (0.00)} & \textbf{3.33 (0.03)} & \textbf{3.68 (0.03)} & \textbf{1.05 (0.00)} & \textbf{3.46 (0.04)} & \textbf{4.39 (0.02)} \\    
           &              & 10000 & \textbf{1.05 (0.00)} & \textbf{3.35 (0.03)} & \textbf{3.75 (0.04)} & \textbf{1.05 (0.00)} & \textbf{3.50 (0.03)} & \textbf{4.36 (0.05)} \\ \cline{2-9}      
           & GRF          &  500  & \textbf{1.30 (0.01)} & \textbf{1.65 (0.02)} & \textbf{2.10 (0.03)} & \textbf{1.12 (0.01)} & \textbf{1.55 (0.02)} & \textbf{2.06 (0.02)} \\    
           &              & 1000  & \textbf{1.17 (0.01)} & \textbf{1.29 (0.01)} & \textbf{1.59 (0.02)} & \textbf{1.08 (0.01)} & \textbf{1.25 (0.01)} & \textbf{1.74 (0.01)} \\    
           &              & 5000  & \textbf{1.04 (0.00)} & 0.99 (0.01)          & 1.00 (0.01)          & \textbf{1.03 (0.00)} & \textbf{1.03 (0.00)} & \textbf{1.20 (0.00)} \\    
           &              & 10000 & \textbf{1.03 (0.00)} & 0.98 (0.00)          & 0.94 (0.01)          & \textbf{1.02 (0.00)} & \textbf{1.02 (0.00)} & \textbf{1.09 (0.00)} \\ \hline

Friedman & R.F.         &  500  &  0.69 (0.01)          & \textbf{1.41 (0.02)} & \textbf{1.51 (0.01)} & 0.76 (0.01)          & \textbf{1.43 (0.02)} & \textbf{1.54 (0.02)} \\       
         &              & 1000  &  0.59 (0.00)          & \textbf{1.34 (0.02)} & \textbf{1.43 (0.02)} & 0.71 (0.00)          & \textbf{1.49 (0.01)} & \textbf{1.61 (0.02)} \\       
         &              & 5000  &  0.41 (0.00)          & 0.89 (0.00)          & \textbf{1.06 (0.01)} & 0.58 (0.00)          & \textbf{1.21 (0.01)} & \textbf{1.41 (0.01)} \\      
         &              & 10000 &  0.35 (0.00)          & 0.78 (0.01)          & 0.93 (0.01)          & 0.50 (0.00)          & \textbf{1.09 (0.01)} & \textbf{1.23 (0.01)} \\ \cline{2-9}          
         & Reduced R.F. &  500  &  \textbf{1.02 (0.01)} & \textbf{1.77 (0.02)} & \textbf{1.79 (0.02)} & \textbf{1.12 (0.01)} & \textbf{1.80 (0.02)} & \textbf{1.83 (0.02)} \\       
         &              & 1000  &  \textbf{1.04 (0.01)} & \textbf{1.92 (0.02)} & \textbf{1.90 (0.02)} & \textbf{1.25 (0.01)} & \textbf{2.13 (0.02)} & \textbf{2.14 (0.02)} \\       
         &              & 5000  &  \textbf{1.06 (0.00)} & \textbf{1.88 (0.01)} & \textbf{1.97 (0.02)} & \textbf{1.50 (0.01)} & \textbf{2.55 (0.02)} & \textbf{2.62 (0.02)} \\       
         &              & 10000 &  \textbf{1.06 (0.00)} & \textbf{1.91 (0.02)} & \textbf{1.99 (0.02)} & \textbf{1.53 (0.01)} & \textbf{2.68 (0.02)} & \textbf{2.61 (0.03)} \\ \cline{2-9}      
         & GRF          &  500  &  \textbf{1.04 (0.01)} & \textbf{1.24 (0.01)} & \textbf{1.37 (0.01)} & \textbf{1.15 (0.01)} & \textbf{1.26 (0.01)} & \textbf{1.41 (0.02)} \\       
         &              & 1000  &  0.88 (0.01)          & \textbf{1.09 (0.01)} & \textbf{1.18 (0.01)} & \textbf{1.06 (0.01)} & \textbf{1.22 (0.01)} & \textbf{1.33 (0.01)} \\       
         &              & 5000  &  0.57 (0.00)          & 0.67 (0.00)          & 0.76 (0.01)          & 0.80 (0.00)          & 0.90 (0.01)          & \textbf{1.02 (0.01)} \\      
         &              & 10000 &  0.46 (0.00)          & 0.57 (0.01)          & 0.66 (0.00)          & 0.66 (0.00)          & 0.80 (0.01)          & 0.87 (0.01)          \\ \hline

Linear   & R.F.         &  500  & 0.73 (0.01)          & \textbf{1.26 (0.01)} & \textbf{1.31 (0.02)} & 0.77 (0.01)          & \textbf{1.31 (0.01)} & \textbf{1.40 (0.02)} \\    
         &              & 1000  & 0.65 (0.01)          & \textbf{1.18 (0.01)} & \textbf{1.25 (0.02)} & 0.73 (0.01)          & \textbf{1.38 (0.01)} & \textbf{1.43 (0.02)} \\    
         &              & 5000  & 0.52 (0.00)          & 0.84 (0.01)          & 0.97 (0.01)          & 0.72 (0.00)          & \textbf{1.17 (0.01)} & \textbf{1.30 (0.02)} \\    
         &              & 10000 & 0.46 (0.00)          & 0.70 (0.01)          & 0.81 (0.01)          & 0.68 (0.00)          & \textbf{1.02 (0.01)} & \textbf{1.11 (0.01)} \\ \cline{2-9}     
         & Reduced R.F. &  500  & \textbf{1.11 (0.01)} & \textbf{1.61 (0.01)} & \textbf{1.62 (0.01)} & \textbf{1.17 (0.01)} & \textbf{1.68 (0.01)} & \textbf{1.73 (0.02)} \\    
         &              & 1000  & \textbf{1.14 (0.01)} & \textbf{1.78 (0.01)} & \textbf{1.73 (0.02)} & \textbf{1.28 (0.01)} & \textbf{2.07 (0.02)} & \textbf{1.98 (0.02)} \\    
         &              & 5000  & \textbf{1.25 (0.01)} & \textbf{1.80 (0.02)} & \textbf{1.84 (0.02)} & \textbf{1.71 (0.01)} & \textbf{2.50 (0.03)} & \textbf{2.45 (0.03)} \\    
         &              & 10000 & \textbf{1.24 (0.01)} & \textbf{1.77 (0.02)} & \textbf{1.77 (0.01)} & \textbf{1.82 (0.01)} & \textbf{2.58 (0.03)} & \textbf{2.44 (0.02)} \\ \cline{2-9}      
         & GRF          &  500  & \textbf{1.08 (0.01)} & \textbf{1.03 (0.01)} & \textbf{1.15 (0.02)} & \textbf{1.15 (0.01)} & \textbf{1.08 (0.01)} & \textbf{1.23 (0.02)} \\    
         &              & 1000  & 0.91 (0.01)          & 0.88 (0.01)          & 1.00 (0.01)          & \textbf{1.02 (0.01)} & \textbf{1.03 (0.01)} & \textbf{1.14 (0.01)} \\    
         &              & 5000  & 0.67 (0.00)          & 0.55 (0.00)          & 0.65 (0.01)          & 0.91 (0.00)          & 0.77 (0.01)          & 0.86 (0.01)          \\    
         &              & 10000 & 0.56 (0.00)          & 0.45 (0.00)          & 0.52 (0.00)          & 0.82 (0.00)          & 0.65 (0.01)          & 0.72 (0.00)          \\ \hline

\end{tabular} 
\end{table}

\begin{table}[h]
\centering
\caption{Mean and standard error of processing time ratios comparing each baseline method to \RanBu{} and \DiNo{} across different scenarios, methods, training sample sizes, and numbers of noise variables. 
Ratios greater than 1 indicate longer processing times relative to the proposed methods. 
Results highlight the substantial computational advantage of \RanBu{}, which scales more favorably with sample size and noise, while \DiNo{} also remains competitive compared to Generalized Random Forests (GRF).}
\label{tab:eqm:simulation:time}
\scriptsize 
\begin{tabular}{lllcccccc}
\toprule
\multirow{2}{*}{Scenario} & \multirow{2}{*}{Method} & \multirow{2}{*}{$n_{\text{train}}$} & \multicolumn{3}{c}{\DiNo} & \multicolumn{3}{c}{\RanBu} \\
\cmidrule(lr){4-6} \cmidrule(lr){7-9}
 &  & & Noise 0 & Noise 50 & Noise 100 & Noise 0 & Noise 50 & Noise 100 \\
\midrule

rectangular & R.F.         &  500  &  0.14 (0.00)           & 0.50 (0.01)           & 0.69 (0.01)           & \textbf{9.47 (0.24) }  & \textbf{12.96 (0.22)}  & \textbf{12.49 (0.08)}  \\    
           &              & 1000  &  0.35 (0.01)           & \textbf{1.17 (0.02) } & \textbf{1.66 (0.01) } & \textbf{19.33 (0.63)}  & \textbf{21.16 (0.33)}  & \textbf{19.52 (0.10)}  \\    
           &              & 5000  &  \textbf{2.56 (0.05) } & \textbf{7.89 (0.13) } & \textbf{10.30 (0.16)} & \textbf{43.87 (0.90)}  & \textbf{46.88 (0.62)}  & \textbf{42.25 (0.70)}  \\    
           &              & 10000 &  \textbf{5.98 (0.18) } & \textbf{15.29 (0.39)} & \textbf{17.84 (0.42)} & \textbf{60.44 (1.78)}  & \textbf{53.96 (0.87)}  & \textbf{52.59 (0.45)}  \\ \cline{2-9}     
           & Reduced R.F. &  500  &  0.01 (0.00)           & 0.02 (0.00)           & 0.04 (0.00)           & 0.53 (0.01)            & 0.63 (0.00)            & 0.65 (0.00)            \\    
           &              & 1000  &  0.01 (0.00)           & 0.04 (0.00)           & 0.06 (0.00)           & 0.61 (0.00)            & 0.68 (0.01)            & 0.68 (0.00)            \\    
           &              & 5000  &  0.05 (0.00)           & 0.13 (0.00)           & 0.18 (0.00)           & 0.78 (0.01)            & 0.77 (0.00)            & 0.73 (0.01)            \\    
           &              & 10000 &  0.09 (0.01)           & 0.22 (0.01)           & 0.25 (0.01)           & 0.80 (0.00)            & 0.76 (0.00)            & 0.73 (0.00)            \\ \cline{2-9}      
           & GRF          &  500  &  0.60 (0.01)           & \textbf{3.10 (0.06) } & \textbf{3.52 (0.05) } & \textbf{41.22 (0.89)}  & \textbf{79.59 (0.75) } & \textbf{63.78 (0.34)}  \\    
           &              & 1000  &  \textbf{1.31 (0.02)  }& \textbf{6.45 (0.05) } & \textbf{7.45 (0.05) } & \textbf{72.15 (1.01)}  & \textbf{116.64 (1.26)} & \textbf{87.86 (0.33)}  \\    
           &              & 5000  &  \textbf{7.47 (0.12) } & \textbf{33.75 (0.56)} & \textbf{34.81 (0.41)} & \textbf{128.57 (2.70)} & \textbf{200.38 (2.40)} & \textbf{142.78 (2.01)} \\    
           &              & 10000 &  \textbf{17.11 (0.37)} & \textbf{60.96 (1.61)} & \textbf{53.23 (1.23)} & \textbf{173.05 (3.98)} & \textbf{214.72 (3.28)} & \textbf{156.90 (1.17)} \\ \hline

Friedman & R.F.         &  500   & 0.15 (0.00)           & 0.54 (0.01)           & 0.76 (0.02)           & \textbf{9.39 (0.24) }  & \textbf{12.44 (0.04)}  & \textbf{12.38 (0.23)} \\       
         &              & 1000   & 0.35 (0.00)           & \textbf{1.24 (0.02) } & \textbf{1.70 (0.01) } & \textbf{16.05 (0.13)}  & \textbf{20.67 (0.31)}  & \textbf{19.35 (0.09)} \\       
         &              & 5000   & \textbf{2.19 (0.04) } & \textbf{6.90 (0.12) } & \textbf{10.00 (0.18)} & \textbf{37.57 (0.91)}  & \textbf{44.03 (0.56)}  & \textbf{41.17 (0.95)} \\      
         &              & 10000  & \textbf{5.17 (0.15) } & \textbf{14.10 (0.32)} & \textbf{19.25 (0.40)} & \textbf{48.31 (0.78)}  & \textbf{40.62 (2.00)}  & \textbf{50.81 (0.62)} \\ \cline{2-9}          
         & Reduced R.F. &  500   & 0.01 (0.00)           & 0.03 (0.00)           & 0.04 (0.00)           & 0.51 (0.01)            & 0.63 (0.00)            & 0.65 (0.01)           \\       
         &              & 1000   & 0.01 (0.00)           & 0.04 (0.00)           & 0.06 (0.00)           & 0.62 (0.00)            & 0.68 (0.01)            & 0.68 (0.00)           \\       
         &              & 5000   & 0.04 (0.00)           & 0.12 (0.00)           & 0.18 (0.01)           & 0.71 (0.02)            & 0.73 (0.01)            & 0.73 (0.01)           \\       
         &              & 10000  & 0.08 (0.00)           & 0.29 (0.02)           & 0.28 (0.01)           & 0.76 (0.01)            & 0.76 (0.02)            & 0.73 (0.01)           \\ \cline{2-9}      
         & GRF          &  500   & 0.74 (0.01)           & \textbf{3.42 (0.06) } & \textbf{3.96 (0.06) } & \textbf{44.89 (0.84)}  & \textbf{78.96 (0.28)}  & \textbf{64.19 (0.92)} \\       
         &              & 1000   & \textbf{1.45 (0.01) } & \textbf{6.98 (0.05) } & \textbf{7.83 (0.05) } & \textbf{67.28 (0.44)}  & \textbf{116.25 (1.07}) & \textbf{89.19 (0.36)} \\       
         &              & 5000   & \textbf{7.51 (0.14) } & \textbf{30.69 (0.52)} & \textbf{34.58 (0.60)} & \textbf{128.84 (3.03}) & \textbf{195.57 (2.08}) & \textbf{142.36 (3.22} \\      
         &              & 10000  & \textbf{17.74 (0.41)} & \textbf{57.03 (1.28)} & \textbf{60.05 (1.24)} & \textbf{166.14 (1.49}) & \textbf{164.60 (8.14}) & \textbf{158.49 (1.91} \\ \hline

Linear   & R.F.         &  500  & 0.17 (0.00)           & 0.60 (0.01)           & 0.80 (0.02)           & \textbf{9.19 (0.13) }  & \textbf{12.89 (0.18)}  & \textbf{12.49 (0.06)}  \\    
         &              & 1000  & 0.36 (0.01)           & \textbf{1.37 (0.03)}  & \textbf{1.72 (0.02) } & \textbf{16.49 (0.49)}  & \textbf{22.02 (0.40)}  & \textbf{19.11 (0.26)}  \\    
         &              & 5000  & \textbf{2.21 (0.04) } & \textbf{7.72 (0.16)}  & \textbf{11.27 (0.13)} & \textbf{37.60 (0.49)}  & \textbf{44.50 (0.53)}  & \textbf{43.69 (0.34)}  \\    
         &              & 10000 & \textbf{5.01 (0.14) } & \textbf{5.41 (0.34) } & \textbf{19.18 (0.42)} & \textbf{47.84 (0.90)}  & \textbf{52.11 (0.83)}  & \textbf{51.37 (0.54)}  \\ \cline{2-9}     
         & Reduced R.F. &  500  & 0.01 (0.00)           & 0.03 (0.00)           & 0.04 (0.00)           & 0.55 (0.00)            & 0.63 (0.00)            & 0.64 (0.00)            \\    
         &              & 1000  & 0.01 (0.00)           & 0.04 (0.00)           & 0.06 (0.00)           & 0.61 (0.01)            & 0.69 (0.00)            & 0.66 (0.01)            \\    
         &              & 5000  & 0.04 (0.00)           & 0.13 (0.00)           & 0.19 (0.00)           & 0.74 (0.01)            & 0.74 (0.01)            & 0.73 (0.00)            \\    
         &              & 10000 & 0.08 (0.00)           & 0.22 (0.00)           & 0.27 (0.01)           & 0.75 (0.01)            & 0.75 (0.01)            & 0.73 (0.00)            \\ \cline{2-9}      
         & GRF          &  500  & 0.76 (0.01)           & \textbf{3.70 (0.06)}  & \textbf{4.21 (0.09) } & \textbf{40.64 (0.21)}  & \textbf{79.95 (0.30)}  & \textbf{65.32 (0.47)}  \\    
         &              & 1000  & \textbf{1.42 (0.01) } & \textbf{7.34 (0.07)}  & \textbf{7.93 (0.08) } & \textbf{64.93 (0.80)}  & \textbf{118.08 (0.40)} & \textbf{87.86 (1.18)}  \\    
         &              & 5000  & \textbf{7.66 (0.11) } & \textbf{3.24 (0.63)}  & \textbf{37.97 (0.34)} & \textbf{130.45 (1.69)} & \textbf{191.51 (1.69)} & \textbf{147.39 (0.99)} \\    
         &              & 10000 & \textbf{17.02 (0.45)} & 0.92 (1.14)           & \textbf{56.73 (1.16)} & \textbf{162.20 (2.28)} & \textbf{206.29 (2.85)} & \textbf{152.18 (1.51)} \\ \hline

\end{tabular}
\end{table}

\begin{table}[h]
\centering
\caption{Mean and standard error of pinball loss ratios comparing each baseline method to \RanBu{} and \DiNo{} across different scenarios, methods, training sample sizes, and numbers of noise variables. 
Ratios greater than 1 indicate better predictive performance of the proposed methods. 
Results show that both \RanBu{} and \DiNo{} become increasingly advantageous in the presence of noise and at larger training sizes, with \RanBu{} generally delivering the most consistent improvements.}

\label{tab:pinball:simulation}
\scriptsize 
\begin{tabular}{lllcccccc}
\toprule
\multirow{2}{*}{Scenario} & \multirow{2}{*}{Method} & \multirow{2}{*}{$n_{\text{train}}$} & \multicolumn{3}{c}{\DiNo} & \multicolumn{3}{c}{\RanBu} \\
\cmidrule(lr){4-6} \cmidrule(lr){7-9}
 &  & & Noise 0 & Noise 50 & Noise 100 & Noise 0 & Noise 50 & Noise 100 \\
\midrule

rectangular & R.F.          &   500 & 0.797 (0.007) &  1.111 (0.017)  &  1.151 (0.021)  &  0.925 (0.002)  &  1.381 (0.008)  &  1.503 (0.010)  \\       
           &               &  1000 & 0.783 (0.006) &  1.056 (0.011)  &  1.139 (0.013)  &  1.001 (0.002)  &  1.380 (0.006)  &  1.558 (0.008)  \\       
           &               &  5000 & 0.770 (0.006) &  0.854 (0.007)  &  0.748 (0.017)  &  1.052 (0.000)  &  1.214 (0.003)  &  1.371 (0.002)  \\       
           &               & 10000 & 0.738 (0.003) &  0.810 (0.004)  &  0.868 (0.009)  &  1.050 (0.001)  &  1.144 (0.002)  &  1.282 (0.003)  \\   \cline{2-9}         
           & Reduced R.F.  &   500 & 0.800 (0.006) &  1.302 (0.021)  &  1.304 (0.022)  &  0.928 (0.003)  &  1.618 (0.010)  &  1.705 (0.010)  \\       
           &               &  1000 & 0.776 (0.007) &  1.347 (0.015)  &  1.404 (0.015)  &  0.992 (0.003)  &  1.760 (0.012)  &  1.921 (0.013)  \\       
           &               &  5000 & 0.767 (0.005) &  1.389 (0.015)  &  1.260 (0.018)  &  1.049 (0.002)  &  1.976 (0.017)  &  2.327 (0.017)  \\       
           &               & 10000 & 0.740 (0.003) &  1.386 (0.010)  &  1.476 (0.020)  &  1.052 (0.002)  &  1.958 (0.012)  &  2.176 (0.011)  \\   \cline{2-9}           
           & GRF           &   500 & 1.372 (0.013) &  1.899 (0.029)  &  1.742 (0.033)  &  1.592 (0.009)  &  2.360 (0.014)  &  2.272 (0.015)  \\       
           &               &  1000 & 1.245 (0.012) &  2.006 (0.022)  &  1.913 (0.022)  &  1.593 (0.008)  &  2.620 (0.015)  &  2.616 (0.015)  \\       
           &               &  5000 & 0.956 (0.007) &  1.595 (0.017)  &  1.308 (0.030)  &  1.306 (0.002)  &  2.267 (0.010)  &  2.397 (0.005)  \\       
           &               & 10000 & 0.796 (0.004) &  1.305 (0.007)  &  1.382 (0.016)  &  1.132 (0.002)  &  1.844 (0.007)  &  2.040 (0.008)  \\   \hline
           
Friedman   & R.F.          &   500 & 0.684 (0.008) &  0.877 (0.013)  &  0.838 (0.012)  &  0.813 (0.003)  &  1.106 (0.004)  &  1.164 (0.007)  \\       
           &               &  1000 & 0.663 (0.002) &  0.847 (0.011)  &  0.782 (0.009)  &  0.816 (0.001)  &  1.181 (0.006)  &  1.210 (0.005)  \\       
           &               &  5000 & 0.533 (0.004) &  0.714 (0.007)  &  0.740 (0.008)  &  0.793 (0.002)  &  1.125 (0.005)  &  1.203 (0.006)  \\       
           &               & 10000 & 0.480 (0.003) &  0.657 (0.007)  &  0.694 (0.003)  &  0.752 (0.002)  &  1.077 (0.005)  &  1.134 (0.005)  \\   \cline{2-9}             
           & Reduced R.F.  &   500 & 0.848 (0.011) &  0.998 (0.015)  &  0.924 (0.012)  &  1.007 (0.004)  &  1.259 (0.006)  &  1.285 (0.006)  \\       
           &               &  1000 & 0.878 (0.003) &  1.022 (0.013)  &  0.926 (0.010)  &  1.081 (0.002)  &  1.424 (0.009)  &  1.435 (0.008)  \\       
           &               &  5000 & 0.858 (0.007) &  1.036 (0.012)  &  1.011 (0.010)  &  1.277 (0.005)  &  1.633 (0.010)  &  1.644 (0.009)  \\       
           &               & 10000 & 0.837 (0.005) &  1.017 (0.010)  &  1.035 (0.007)  &  1.311 (0.005)  &  1.668 (0.009)  &  1.690 (0.005)  \\    \cline{2-9}            
           & GRF           &   500 & 0.969 (0.010) &  1.166 (0.017)  &  1.077 (0.015)  &  1.153 (0.005)  &  1.471 (0.007)  &  1.497 (0.011)  \\       
           &               &  1000 & 0.964 (0.005) &  1.150 (0.016)  &  1.030 (0.012)  &  1.186 (0.003)  &  1.602 (0.010)  &  1.594 (0.008)  \\       
           &               &  5000 & 0.772 (0.006) &  0.911 (0.010)  &  0.939 (0.011)  &  1.148 (0.004)  &  1.436 (0.008)  &  1.526 (0.009)  \\       
           &               & 10000 & 0.706 (0.005) &  0.823 (0.009)  &  0.870 (0.005)  &  1.106 (0.004)  &  1.349 (0.007)  &  1.420 (0.006)  \\   \hline
           
Linear     & R.F.          &   500 & 0.679 (0.006) &  0.842 (0.015)  &  0.863 (0.017)  &  0.802 (0.003)  &  1.083 (0.005)  &  1.143 (0.007)  \\       
           &               &  1000 & 0.643 (0.009) &  0.804 (0.012)  &  0.841 (0.010)  &  0.820 (0.002)  &  1.113 (0.007)  &  1.178 (0.006)  \\       
           &               &  5000 & 0.561 (0.006) &  0.695 (0.007)  &  0.741 (0.008)  &  0.848 (0.002)  &  1.093 (0.005)  &  1.157 (0.007)  \\       
           &               & 10000 & 0.519 (0.004) &  0.635 (0.005)  &  0.690 (0.005)  &  0.842 (0.002)  &  1.053 (0.006)  &  1.102 (0.006)  \\   \cline{2-9}             
           & Reduced R.F.  &   500 & 0.847 (0.009) &  0.983 (0.018)  &  0.962 (0.018)  &  1.001 (0.005)  &  1.265 (0.007)  &  1.276 (0.008)  \\       
           &               &  1000 & 0.864 (0.012) &  0.990 (0.014)  &  1.007 (0.011)  &  1.101 (0.005)  &  1.372 (0.010)  &  1.411 (0.008)  \\       
           &               &  5000 & 0.879 (0.009) &  1.008 (0.010)  &  1.021 (0.010)  &  1.329 (0.006)  &  1.585 (0.008)  &  1.594 (0.008)  \\       
           &               & 10000 & 0.866 (0.007) &  0.981 (0.008)  &  1.021 (0.008)  &  1.406 (0.004)  &  1.628 (0.007)  &  1.630 (0.007)  \\   \cline{2-9}             
           & GRF           &   500 & 1.020 (0.010) &  1.110 (0.021)  &  1.092 (0.023)  &  1.205 (0.005)  &  1.426 (0.008)  &  1.446 (0.011)  \\       
           &               &  1000 & 0.969 (0.014) &  1.061 (0.017)  &  1.075 (0.013)  &  1.236 (0.006)  &  1.468 (0.010)  &  1.505 (0.008)  \\       
           &               &  5000 & 0.843 (0.010) &  0.897 (0.010)  &  0.945 (0.011)  &  1.273 (0.006)  &  1.411 (0.008)  &  1.474 (0.010)  \\       
           &               & 10000 & 0.774 (0.007) &  0.813 (0.008)  &  0.881 (0.007)  &  1.256 (0.004)  &  1.349 (0.009)  &  1.406 (0.008)  \\   \hline

\end{tabular}
\end{table}

\begin{table}[h]
\centering
\caption{Mean and standard error of processing time ratios comparing each baseline method to \RanBu{} and \DiNo{} in quantile regression, 
across different scenarios, training sample sizes, and levels of noise variables. 
Ratios greater than 1 indicate longer processing times compared to the proposed methods. 
\RanBu{} is consistently the most computationally efficient among the full-forest approaches, 
though the Reduced R.F.\ remains faster overall by construction. 
\DiNo{} requires somewhat higher costs than \RanBu{}, but remains substantially lighter than GRF, 
which exhibits the steepest and least stable growth in runtime.}

\label{tab:pinball:simulation:time}
\scriptsize  
\begin{tabular}{lllcccccc}
\toprule
\multirow{2}{*}{Scenario} & \multirow{2}{*}{Method} & \multirow{2}{*}{$n_{\text{train}}$} & \multicolumn{3}{c}{\DiNo} & \multicolumn{3}{c}{\RanBu} \\
\cmidrule(lr){4-6} \cmidrule(lr){7-9}
 &  & & Noise 0 & Noise 50 & Noise 100 & Noise 0 & Noise 50 & Noise 100 \\
\midrule

rectangular & R.F.         &   500 & 0.175 (0.003)   & 0.582 (0.015)    & 0.691 (0.017)     & 9.474 (0.252)    & 12.380 (0.240)    & 11.371 (0.208)   \\
           &              &  1000 & 0.396 (0.007)   & 1.266 (0.021)    & 1.691 (0.028)     & 14.264 (0.240)   & 17.648 (0.366)    & 17.333 (0.324)   \\
           &              &  5000 & 2.875 (0.036)   & 7.633 (0.147)    & 11.775 (0.137)    & 26.216 (0.344)   & 32.598 (0.890)    & 36.547 (0.397)   \\
           &              & 10000 & 5.151 (0.065)   & 14.759 (0.254)   & 18.121 (0.296)    & 25.797 (0.713)   & 41.361 (0.316)    & 40.808 (0.494)   \\  \cline{2-9}      
           & Reduced R.F. &   500 & 0.014 (0.000)   & 0.037 (0.001)    & 0.047 (0.001)     & 0.750 (0.014)    & 0.778 (0.005)     & 0.771 (0.005)    \\
           &              &  1000 & 0.024 (0.002)   & 0.057 (0.001)    & 0.076 (0.001)     & 0.833 (0.065)    & 0.788 (0.007)     & 0.776 (0.005)    \\
           &              &  5000 & 0.086 (0.001)   & 0.183 (0.004)    & 0.256 (0.004)     & 0.780 (0.003)    & 0.770 (0.015)     & 0.792 (0.004)    \\
           &              & 10000 & 0.170 (0.004)   & 0.286 (0.004)    & 0.356 (0.007)     & 0.826 (0.005)    & 0.801 (0.002)     & 0.797 (0.004)    \\  \cline{2-9}      
           & GRF          &   500 & 0.565 (0.008)   & 6.124 (0.133)    & 6.028 (0.134)     & 30.706 (0.835)   & 130.904 (2.837)   & 99.405 (1.628)   \\
           &              &  1000 & 1.184 (0.015)   & 12.409 (0.258)   & 13.040 (0.207)    & 42.747 (0.697)   & 172.787 (4.044)   & 133.579 (2.339)  \\ 
           &              &  5000 & 6.290 (0.082)   & 56.826 (0.930)   & 63.822 (0.808)    & 57.349 (0.774)   & 243.852 (6.984)   & 198.309 (2.729)  \\ 
           &              & 10000 & 10.350 (0.142)  & 80.527 (1.207)   & 79.711 (1.272)    & 51.974 (1.570)   & 227.242 (3.923)   & 180.181 (3.019)  \\   \hline
           
Friedman   & R.F.          &   500 & 0.211 ( 0.006)  & 0.626 ( 0.015)   & 0.723 ( 0.021)    & 9.684 ( 0.246)   & 12.615 ( 0.293)   & 11.244 ( 0.196)  \\ 
           &               &  1000 & 0.437 ( 0.005)  & 1.368 ( 0.017)   & 1.707 ( 0.020)    & 13.734 ( 0.254)  & 16.778 ( 0.304)   & 16.038 ( 0.213)  \\ 
           &               &  5000 & 2.708 ( 0.028)  & 8.060 ( 0.075)   & 10.875 ( 0.076)   & 25.350 ( 0.338)  & 33.244 ( 0.347)   & 32.910 ( 0.315)  \\ 
           &               & 10000 & 5.370 ( 0.103)  & 12.414 ( 0.350)  & 15.987 ( 0.218)   & 30.068 ( 0.706)  & 38.440 ( 1.119)   & 32.274 ( 0.397)  \\   \cline{2-9}      
           & Reduced R.F.  &   500 & 0.016 ( 0.000)  & 0.038 ( 0.001)   & 0.050 ( 0.002)    & 0.734 ( 0.007)   & 0.762 ( 0.009)    & 0.767 ( 0.007)   \\
           &               &  1000 & 0.024 ( 0.000)  & 0.065 ( 0.001)   & 0.084 ( 0.002)    & 0.754 ( 0.006)   & 0.782 ( 0.008)    & 0.780 ( 0.007)   \\
           &               &  5000 & 0.082 ( 0.001)  & 0.195 ( 0.002)   & 0.257 ( 0.003)    & 0.763 ( 0.004)   & 0.801 ( 0.002)    & 0.776 ( 0.004)   \\
           &               & 10000 & 0.135 ( 0.003)  & 0.270 ( 0.012)   & 0.411 ( 0.004)    & 0.753 ( 0.015)   & 0.797 ( 0.006)    & 0.829 ( 0.002)   \\   \cline{2-9}      
           & GRF           &   500 & 0.818 ( 0.012)  & 5.897 ( 0.100)   & 6.357 ( 0.168)    & 37.614 ( 0.601)  & 119.340 ( 2.426)  & 99.070 ( 1.592)  \\ 
           &               &  1000 & 1.697 ( 0.019)  & 12.603 ( 0.119)  & 13.772 ( 0.169)   & 53.327 ( 0.979)  & 154.814 ( 2.698)  & 129.382 ( 1.697) \\ 
           &               &  5000 & 9.352 ( 0.120)  & 57.967 ( 0.485)  & 61.967 ( 0.416)   & 87.753 ( 1.626)  & 239.399 ( 2.877)  & 187.454 ( 1.587) \\ 
           &               & 10000 & 16.497 ( 0.239) & 76.822 ( 2.101)  & 84.289 (10.197)   & 93.233 ( 2.571)  & 237.335 ( 6.761)  & 171.088 (21.041) \\   \hline
Linear     & R.F.          &   500 & 0.208 (0.005)   & 0.550 (0.014)    & 0.807 (0.018)     & 10.051 (0.302)   & 11.772 (0.394)    & 11.778 (0.201)   \\
           &               &  1000 & 0.426 (0.008)   & 1.308 (0.020)    & 1.851 (0.026)     & 14.636 (0.292)   & 18.528 (0.334)    & 17.788 (0.224)   \\
           &               &  5000 & 2.414 (0.037)   & 8.227 (0.131)    & 10.516 (0.231)    & 24.477 (0.334)   & 34.890 (0.598)    & 33.569 (0.734)   \\
           &               & 10000 & 4.372 (0.103)   & 14.407 (0.248)   & 19.158 (0.253)    & 27.219 (0.724)   & 41.270 (0.673)    & 40.046 (0.590)   \\   \cline{2-9}      
           & Reduced R.F.  &   500 & 0.016 (0.000)   & 0.035 (0.001)    & 0.052 (0.001)     & 0.759 (0.008)    & 0.740 (0.019)     & 0.763 (0.003)    \\
           &               &  1000 & 0.022 (0.000)   & 0.055 (0.001)    & 0.081 (0.002)     & 0.765 (0.005)    & 0.775 (0.005)     & 0.774 (0.005)    \\
           &               &  5000 & 0.075 (0.001)   & 0.190 (0.005)    & 0.254 (0.005)     & 0.760 (0.004)    & 0.796 (0.004)     & 0.803 (0.004)    \\
           &               & 10000 & 0.128 (0.007)   & 0.283 (0.006)    & 0.374 (0.004)     & 0.762 (0.013)    & 0.802 (0.003)     & 0.779 (0.006)    \\   \cline{2-9}      
           & GRF           &   500 & 0.811 (0.012)   & 5.740 (0.146)    & 6.955 (0.130)     & 38.929 (0.638)   & 123.138 (4.153)   & 101.448 (0.864)  \\ 
           &               &  1000 & 1.653 (0.019)   & 12.847 (0.209)   & 14.635 (0.206)    & 56.885 (0.904)   & 182.254 (3.754)   & 140.972 (2.255)  \\ 
           &               &  5000 & 9.414 (0.103)   & 58.792 (1.261)   & 58.359 (0.854)    & 95.724 (1.325)   & 249.215 (5.245)   & 186.385 (3.067)  \\ 
           &               & 10000 & 14.798 (0.350)  & 89.704 (1.688)   & 89.830 (1.224)    & 92.274 (2.700)   & 258.017 (5.573)   & 188.018 (3.181)  \\  \hline


\end{tabular}
\end{table}

\begin{table}[h]
\centering
\caption{\label{tab_descricao_dados} Databases used in the applications of the methods.}
\begin{tabular}{lccl}
\hline
\textbf{data set}              & \textbf{sample size} & \textbf{nº of predictors}  & \textbf{reference}    \\ \hline 
Abalone                        & 4177       & 8                    & \cite{abalone}                    \\        
Adult                          & 48842      & 13                   & \cite{adultdata}                  \\        
Airfoil self-noise             & 1503       & 5                    & \cite{airfoil}                    \\        
Ames Housing                   & 2930       & 80                   & \cite{ameshousing}                \\        
Auto MPG                       & 398        & 8                    & \cite{autompg}                    \\        
Boston                         & 506        & 13                   & \cite{boston}                     \\        
College                        & 717        & 17                   & \cite{pacote::ISLR}               \\        
Communities Crime              & 1994       & 100                  & \cite{communitiescrime}           \\        
Computer Hardware              & 209        & 9                    & \cite{computerhardware}           \\        
Concrete Compressive Strength  & 1030       &  8                   & \cite{concrete}                   \\        
Energy Efficiency              & 768        & 8                    & \cite{energyefficiency}           \\        
Fertil2                        & 4361	     & 26                   & \cite{wooldridge2009introductory} \\        
Forest fires                   & 517        & 12                   & \cite{forestfires}                \\        
Garments worker productivity   & 1197       & 14                   & \cite{garments}                   \\        
injury                         & 7150       & 29                   & \cite{wooldridge2009introductory} \\	      
Online news popularity         & 39644      & 59                   & \cite{onlinenews}                 \\        
QSAR fish toxicity             & 908        & 6                    & \cite{qsar}                       \\        
Real estate valuation          & 414        & 7                    & \cite{realestate}                 \\        
Residential Building           & 372        & 107                  & \cite{residentialbuilding}        \\        
Seoul Bike Sharing             & 8760       & 13                   & \cite{seoulbike}                  \\        
Student (math)                 & 649        & 30                   & \cite{student_math}               \\        
Volkswagem                     & 15157      & 8                    & \cite{pacote::ISLR}               \\        
Wage                           & 3000       & 9                    & \cite{pacote::ISLR}               \\        
Wine Quality                   & 4898       & 11                   & \cite{winequality}                \\        
Yacht hydrodynamics            & 308        & 6                    & \cite{yacht}                      \\ \hline 
\end{tabular}%
\end{table}

\newpage

\newpage

\begin{table}[htbp]
  \centering
\caption{Relative mean squared error (MSE) and standard error for each method, reported as the ratio to the MSE of standard Random Forests (R.F.). 
Each entry is based on the average over 50 replications; values below 1 indicate better performance than R.F. 
\RanBu{} frequently achieves lower error than R.F.\ and stands as the most competitive of the proposed methods. 
\DiNo{} shows stable performance across datasets, though it does not outperform R.F.\ in this metric. 
The Reduced R.F.\ baseline performs well in certain structured problems, while GRF remains a strong but more computationally demanding alternative.}

\label{tab:eqm:application_ratio}
\scriptsize  
  \begin{tabular}{lccccc}
  \hline
  \textbf{data set}      & \textbf{Boosting}    & \textbf{\DiNo{}} & \textbf{GRF}    & \textbf{\RanBu{}}      & \textbf{Reduced R.F.}  \\ \hline 
  
Abalone (n = 4177)                       & 1.193 (0.005)  & 1.184 (0.005) & \textbf{1.009 (0.015)}  & 1.082 (0.004) & 1.178 (0.005) \\
Adult (n = 48842)                        & 1.115 (0.001)  & -             & \textbf{1.002 (0.003)}  & 1.054 (0.001) & 1.190 (0.002) \\
Airfoil self-noise (n = 1503)            & 4.237 (0.070)  & 2.737 (0.038) & \textbf{1.531 (0.034)}  & 1.589 (0.021) & 2.923 (0.040) \\
Ames Housing (n = 2930)                  & 1.259 (0.018)  & 1.355 (0.014) & 1.527 (0.072)  & \textbf{1.159 (0.021)} & 1.562 (0.014) \\
Auto MPG (n = 398)                       & 18.282 (0.638) & \textbf{1.026 (0.014)} & 3.282 (0.148)  & 1.032 (0.021) & 1.140 (0.010) \\
Boston (n = 506)                         & 1.400 (0.031)  & 1.052 (0.027) & 1.684 (0.097)  & \textbf{0.956 (0.033)} & 1.291 (0.017) \\
College (n = 717)                        & 1.152 (0.037)  & \textbf{1.089 (0.026)} & 2.254 (0.469)  & 1.099 (0.027) & 1.152 (0.021) \\
Communities Crime (n = 1994)             & \textbf{1.037 (0.005)}  & 1.074 (0.005) & 1.083 (0.027)  & 1.176 (0.010) & 1.060 (0.003) \\
Computer Hardware (n = 209)              & 5.105 (0.478)  & 1.172 (0.108) & 14.641 (4.299) & 1.334 (0.190) & \textbf{1.120 (0.039)} \\
Concrete Compressive Strength (n = 1030) & 1.710 (0.023)  & 1.525 (0.021) & 1.445 (0.036)  & \textbf{1.136 (0.017)} & 2.222 (0.030) \\
Energy Efficiency (n = 768)              & 3.342 (0.057)  & 1.103 (0.030) & 0.525 (0.038)  & \textbf{0.239 (0.009)} & 1.980 (0.028) \\
Fertil2 (n = 4361)                       & 0.631 (0.011)  & 1.255 (0.018) & \textbf{0.603 (0.015)}  & 0.974 (0.012) & 2.085 (0.026) \\
Forest fires (n = 517)                   & \textbf{0.956 (0.023)}  & 1.259 (0.117) & 3.279 (0.682)  & 1.651 (0.191) & 1.039 (0.026) \\
Garments worker productivity (n = 1197)  & 1.291 (0.014)  & 1.155 (0.010) & 1.159 (0.032)  & \textbf{1.123 (0.012)} & 1.187 (0.009) \\
Injury (n = 7150)                        & 0.164 (0.004)  & 4.637 (0.093) & \textbf{0.015 (0.003)}  & 1.965 (0.040) & 6.230 (0.133) \\
Online news popularity (n = 39644)       & \textbf{0.987 (0.003)}  & -             & 1.230 (0.105)  & 1.051 (0.025) & 0.994 (0.002) \\
QSAR fish toxicity (n = 908)             & 1.129 (0.010)  & 1.155 (0.009) & 1.241 (0.037)  & 1.176 (0.014) & \textbf{1.128 (0.007)} \\
Real estate valuation (n = 414)          & 1.090 (0.020)  & 1.080 (0.013) & 1.358 (0.129)  & 1.224 (0.033) & \textbf{1.062 (0.012)} \\
Residential Building (n = 372)           & \textbf{0.428 (0.024)}  & 1.160 (0.023) & 1.314 (0.183)  & 1.102 (0.026) & 1.257 (0.023) \\
Seoul Bike Sharing (n = 8760)            & 2.911 (0.020)  & 2.198 (0.022) & \textbf{1.159 (0.012)}  & 1.455 (0.008) & 2.627 (0.020) \\
Student - math (n = 649)                 & \textbf{1.044 (0.010)}  & 1.073 (0.012) & 1.070 (0.036)  & 1.245 (0.021) & 1.057 (0.007) \\
Volkswagen (n = 15157)                   & 5.904 (0.117)  & 3.484 (0.095) & 1.433 (0.044)  & \textbf{1.138 (0.020)} & 3.942 (0.074) \\
Wage (n = 3000)                          & 0.986 (0.003)  & 1.004 (0.003) & \textbf{0.984 (0.022)}  & 1.043 (0.004) & 1.002 (0.003) \\
Wine quality (n = 4898)                  & 1.453 (0.008)  & 1.415 (0.007) & \textbf{1.310 (0.013)}  & 1.258 (0.006) & 1.436 (0.007) \\
Yacht hydrodynamics (n = 308)            & \textbf{0.175 (0.012)}  & 1.749 (0.125) & 0.229 (0.025)  & 1.844 (0.125) & 1.591 (0.059) \\ \hline 

\end{tabular}%
\end{table}

\begin{table}[htbp]
  \centering
\caption{Ratio of mean execution time (with standard error) for each method relative to standard Random Forests (R.F.), across 50 replications. 
Ratios below 1 indicate computational gains compared to R.F. 
\RanBu{} is consistently the most computationally efficient among the proposed methods, running faster than R.F.\ across all datasets. 
\DiNo{}, in contrast, is substantially more expensive, often by an order of magnitude, reflecting the additional cost of the MRCA-based distance. 
The Reduced R.F.\ baseline unsurprisingly yields the lowest runtimes overall, while Boosting also offers large computational savings relative to R.F.\ and GRF.}

\label{tab:time:application_ratio}
\scriptsize  
  \begin{tabular}{lccccc}
  \hline
  \textbf{data set}      & \textbf{Boosting}    & \textbf{\DiNo{}} & \textbf{GRF}    & \textbf{\RanBu{}}      & \textbf{Reduced R.F.}  \\ \hline

Abalone (n = 4177)                       & 0.054 (0.001) & 2.263 (0.037)  & 7.057 (0.119)  & 0.040 (0.001) & \textbf{0.025 (0.001)} \\
Adult (n = 48842)                        & 0.065 (0.001) & -              & 56.564 (2.034) & 0.032 (0.001) & \textbf{0.024 (0.000)} \\
Airfoil self-noise (n = 1503)            & \textbf{0.051 (0.001)} & 12.123 (0.265) & 8.897 (0.211)  & 0.104 (0.003) & 0.053 (0.002) \\
Ames Housing (n = 2930)                  & 0.179 (0.004) & 1.675 (0.023)  & 8.848 (0.115)  & 0.050 (0.001) & \textbf{0.035 (0.001)} \\
Auto MPG (n = 398)                       & 0.138 (0.004) & 28.586 (0.507) & 8.012 (0.183)  & 0.228 (0.004) & \textbf{0.091 (0.003)} \\
Boston (n = 506)                         & 0.078 (0.002) & 14.411 (0.231) & 7.707 (0.230)  & 0.147 (0.003) & \textbf{0.070 (0.003)} \\
College (n = 717)                        & 0.085 (0.003) & 7.205 (0.188)  & 7.533 (0.247)  & 0.084 (0.003) & \textbf{0.048 (0.002)} \\
Communities Crime (n = 1994)             & 0.176 (0.004) & 2.319 (0.031)  & 9.155 (0.173)  & 0.055 (0.001) & \textbf{0.037 (0.001)} \\
Computer Hardware (n = 209)              & \textbf{0.118 (0.004)} & 29.098 (0.616) & 6.510 (0.198)  & 0.379 (0.011) & 0.125 (0.006) \\
Concrete Compressive Strength (n = 1030) & 0.062 (0.001) & 13.283 (0.273) & 9.005 (0.231)  & 0.116 (0.004) & \textbf{0.060 (0.003)} \\
Energy Efficiency (n = 768)              & 0.128 (0.005) & 26.016 (0.486) & 12.477 (0.254) & 0.272 (0.007) & \textbf{0.122 (0.005)} \\
Fertil2 (n = 4361)                       & 0.117 (0.006) & 3.731 (0.053)  & 14.437 (0.302) & 0.076 (0.006) & \textbf{0.047 (0.002)} \\
Forest fires (n = 517)                   & 0.123 (0.004) & 15.883 (0.282) & 15.589 (0.292) & 0.179 (0.005) & \textbf{0.079 (0.004)} \\
Garments worker productivity (n = 1197)  & 0.103 (0.003) & 10.051 (0.207) & 13.091 (0.290) & 0.119 (0.004) & \textbf{0.066 (0.004)} \\
Injury (n = 7150)                        & 0.082 (0.004) & 0.998 (0.014)  & 8.563 (0.095)  & 0.036 (0.001) & \textbf{0.026 (0.001)} \\
Online news popularity (n = 39644)       & 0.032 (0.000) & -              & 10.506 (0.132) & 0.013 (0.000) & \textbf{0.010 (0.000)} \\
QSAR fish toxicity (n = 908)             & \textbf{0.048 (0.002)} & 11.694 (0.260) & 7.258 (0.682)  & 0.107 (0.003) & 0.050 (0.002) \\
Real estate valuation (n = 414)          & \textbf{0.070 (0.002)} & 26.090 (0.437) & 5.625 (0.088)  & 0.220 (0.005) & 0.091 (0.004) \\
Residential Building (n = 372)           & 0.378 (0.012) & 12.163 (0.200) & 7.315 (0.135)  & 0.179 (0.005) & \textbf{0.105 (0.004)} \\
Seoul Bike Sharing (n = 8760)            & 0.068 (0.002) & 0.999 (0.013)  & 13.637 (0.197) & 0.037 (0.001) & \textbf{0.026 (0.001)} \\
Student - math (n = 649)                 & 0.287 (0.021) & 21.225 (0.351) & 13.499 (0.296) & \textbf{0.248 (0.016)} & 0.127 (0.016) \\
Volkswagen (n = 15157)                   & 0.057 (0.001) & 0.906 (0.011)  & 17.068 (0.452) & 0.046 (0.002) & \textbf{0.030 (0.001)} \\
Wage (n = 3000)                          & 0.079 (0.003) & 6.319 (0.089)  & 24.218 (0.378) & 0.083 (0.002) & \textbf{0.050 (0.002)} \\
Wine quality (n = 4898)                  & 0.055 (0.002) & 2.162 (0.033)  & 12.915 (0.185) & 0.049 (0.002) & \textbf{0.032 (0.001)} \\
Yacht hydrodynamics (n = 308)            & 0.131 (0.016) & 33.689 (0.682) & 7.889 (0.168)  & 0.336 (0.007) & \textbf{0.115 (0.003)} \\ \hline 

\end{tabular}%
\end{table}

\begin{table}[htbp]
  \centering
\caption{Relative pinball loss (with standard error) for each method, reported as the ratio to the pinball loss of standard Random Forests (R.F.), averaged over 50 replications. Values below 1 indicate better performance than R.F. Overall, R.F.\ is a strong baseline for quantile regression: \RanBu{} shows the most consistent gains when improvements occur (e.g., \emph{Energy Efficiency}) and is otherwise often close to parity with R.F.; \DiNo{} displays stable behavior across datasets, occasionally approaching R.F. but not consistently surpassing it. GRF yields improvements in some tasks (e.g., \emph{Abalone}, \emph{Fertil2}, \emph{Injury}), though with variability across datasets, while Reduced R.F.\ remains close to R.F.\ but seldom provides clear benefits.}

\label{tab:pinball:application_ratio}
\footnotesize  
  \begin{tabular}{lccccc}
  \hline
  \textbf{data set}      & \textbf{\DiNo{}}    & \textbf{GRF} & \textbf{\RanBu{}}        & \textbf{Reduced R.F.}  \\ \hline        
Abalone (n = 4177)                       & 1.085 (0.015)  & \textbf{0.997 (0.048)} & 1.056 (0.013) & 1.114 (0.016) \\ 
Airfoil self-noise (n = 1503)            & 1.578 (0.058)  & 1.650 (0.087) & \textbf{1.248 (0.044)} & 1.691 (0.054) \\ 
Ames Housing (n = 2930)                  & 1.172 (0.028)  & 1.354 (0.088) & \textbf{1.159 (0.032)} & 1.396 (0.030) \\ 
Auto MPG (n = 398)                       & \textbf{1.052 (0.037)} & 1.930 (0.256) & 1.156 (0.071) & 1.117 (0.042) \\ 
Boston (n = 506)                         & \textbf{1.056 (0.084)} & 1.425 (0.219) & 1.093 (0.089) & 1.209 (0.039) \\ 
College (n = 717)                        & \textbf{1.116 (0.061)} & 1.750 (0.452) & 1.144 (0.091) & 1.231 (0.063) \\ 
Communities Crime (n = 1994)             & \textbf{1.052 (0.020)} & 1.061 (0.068) & 1.181 (0.031) & 1.064 (0.016) \\ 
Computer Hardware (n = 209)              & 1.136 (0.186) & 2.628 (1.546) & \textbf{1.089 (0.205)} & 1.202 (0.097) \\ 
Concrete Compressive Strength (n = 1030) & 1.285 (0.050) & 1.470 (0.108) & \textbf{1.128 (0.062)} & 1.628 (0.068) \\ 
Energy Efficiency (n = 768)              & 1.139 (0.094) & 1.810 (0.207) & \textbf{0.540 (0.042)} & 1.643 (0.082) \\ 
Fertil2 (n = 4361)                       & 1.271 (0.087) & \textbf{0.682 (0.069)} & 1.034 (0.065) & 1.856 (0.112) \\ 
Forest fires (n = 517)                   & 1.097 (0.202) & 1.237 (0.784) & 1.305 (0.263) & \textbf{0.992 (0.026)} \\ 
Garments worker productivity (n = 1197)  & 1.154 (0.036) & 1.207 (0.104) & \textbf{1.125 (0.041)} & 1.243 (0.048) \\ 
Injury (n = 7150)                        & 2.913 (0.190) & \textbf{0.235 (0.025)} & 1.798 (0.133) & 3.563 (0.165) \\ 
Online news popularity (n = 39644)       & -             & \textbf{1.001 (0.084)} & 1.027 (0.015) & 1.037 (0.006) \\ 
QSAR fish toxicity (n = 908)             & \textbf{1.108 (0.033)} & 1.108 (0.089) & 1.180 (0.040) & 1.117 (0.031) \\ 
Residential Building (n = 372)           & \textbf{1.088 (0.059)} & 1.271 (0.327) & 1.166 (0.085) & 1.183 (0.054) \\ 
Real estate valuation (n = 414)          & \textbf{1.061 (0.039)} & 1.187 (0.216) & 1.212 (0.074) & 1.061 (0.037) \\ 
Seoul Bike Sharing (n = 8760)            & 1.577 (0.046) & \textbf{1.154 (0.041)} & 1.258 (0.026) & 1.819 (0.060) \\ 
Student - math (n = 649)                 & 1.077 (0.065) & 1.047 (0.096) & 1.283 (0.093) & \textbf{1.060 (0.029)} \\ 
Volkswagen (n = 15157)                   & 2.303 (0.104) & \textbf{1.002 (0.045)} & 1.446 (0.044) & 2.827 (0.103) \\ 
Wage (n = 3000)                          & 1.000 (0.009) & \textbf{0.994 (0.067)} & 1.047 (0.014) & 1.018 (0.016) \\ 
Wine quality (n = 4898)                  & 1.339 (0.030) & 1.244 (0.067) & \textbf{1.228 (0.027)} & 1.398 (0.034) \\ 
Yacht hydrodynamics (n = 308)            & 1.751 (0.286) & 3.288 (0.813) & 1.888 (0.337) & \textbf{1.352 (0.127)}  \\
Slice Localization (n = 53500)           & -             & \textbf{3.903 (0.107)} & 8.931 (0.268) & 26.138 (0.527) \\ \hline

\end{tabular}%
\end{table}

\begin{table}[htbp]
  \centering
\caption{Ratio of mean execution time (with standard error) for each method relative to standard Random Forests (R.F.) in quantile regression, averaged over 50 replications. Ratios below 1 indicate computational gains compared to R.F. As expected, Reduced R.F.\ is the most computationally efficient, consistently attaining the lowest runtimes. \RanBu{} also delivers strong efficiency gains across nearly all datasets, ranking just behind Reduced R.F. \DiNo{} maintains stable runtimes and, while not designed primarily for speed, it remains competitive on several datasets (e.g., \emph{Ames Housing}, \emph{Injury}, \emph{Seoul Bike Sharing}). GRF is generally the most computationally intensive, often requiring substantially more time than R.F.}

\label{tab:time:application_ratio:quantile}
\footnotesize  
  \begin{tabular}{lccccc}
  \hline
  \textbf{data set}      & \textbf{\DiNo{}}    & \textbf{GRF} & \textbf{\RanBu{}}        & \textbf{Reduced R.F.}  \\ \hline   
Abalone (n = 4177)                       & 1.020 (0.131) & 11.409 (1.364) & 0.062 (0.010)  & \textbf{0.047 (0.009)}  \\
Airfoil self-noise (n = 1503)            & 4.091 (0.522) & 7.574 (0.948)  & 0.144 (0.029)  & \textbf{0.099 (0.022)}  \\
Ames Housing (n = 2930)                  & 0.800 (0.069) & 9.235 (0.786)  & 0.072 (0.021)  & \textbf{0.053 (0.019)}  \\
Auto MPG (n = 398)                       & 9.424 (1.053) & 4.411 (0.612)  & 0.239 (0.048)  & \textbf{0.182 (0.037)}  \\
Boston (n = 506)                         & 5.531 (0.356) & 13.786 (1.248) & 0.177 (0.022)  & \textbf{0.137 (0.022)}  \\
College (n = 717)                        & 3.139 (0.220) & 23.857 (1.650) & 0.114 (0.015)  & \textbf{0.087 (0.016)}  \\
Communities Crime (n = 1994)             & 1.097 (0.119) & 16.783 (1.873) & 0.082 (0.009)  & \textbf{0.061 (0.008)}  \\
Computer Hardware (n = 209)              & 8.128 (1.093) & 5.293 (0.463)  & 0.309 (0.032)  & \textbf{0.237 (0.032)}  \\
Concrete Compressive Strength (n = 1030) & 4.819 (0.357) & 12.731 (0.820) & 0.147 (0.024)  & \textbf{0.107 (0.019)}  \\
Energy Efficiency (n = 768)              & 7.048 (0.647) & 8.289 (0.916)  & 0.260 (0.046)  & \textbf{0.194 (0.043)}  \\
Fertil2 (n = 4361)                       & 1.480 (0.200) & 15.253 (2.407) & 0.093 (0.017)  & \textbf{0.073 (0.014)}  \\
Forest fires (n = 517)                   & 5.729 (0.760) & 17.003 (1.495) & 0.213 (0.036)  & \textbf{0.170 (0.037)}  \\
Garments worker productivity (n = 1197)  & 3.567 (0.280) & 13.383 (1.308) & 0.161 (0.026)  & \textbf{0.119 (0.026)}  \\
Injury (n = 7150)                        & 0.480 (0.029) & 16.514 (1.372) & 0.060 (0.012)  & \textbf{0.043 (0.012)}  \\
Online news popularity (n = 39644)       & -             & 9.275 (1.353)  & 0.019 (0.001)  & \textbf{0.014 (0.001)}  \\
QSAR fish toxicity (n = 908)             & 3.876 (0.213) & 11.823 (0.743) & 0.146 (0.017)  & \textbf{0.106 (0.017)}  \\
Residential Building (n = 372)           & 4.383 (0.288) & 15.271 (1.047) & 0.205 (0.015)  & \textbf{0.157 (0.016)}  \\
Real estate valuation (n = 414)          & 7.459 (0.561) & 9.773 (0.678)  & 0.228 (0.014)  & \textbf{0.174 (0.010)}  \\
Seoul Bike Sharing (n = 8760)            & 0.486 (0.038) & 19.255 (1.313) & 0.069 (0.015)  & \textbf{0.044 (0.012)}  \\
Student - math (n = 649)                 & 6.292 (0.291) & 11.578 (1.074) & 0.231 (0.013)  & \textbf{0.175 (0.010)}  \\
Volkswagen (n = 15157)                   & 0.429 (0.027) & 19.448 (0.860) & 0.069 (0.008)  & \textbf{0.051 (0.007)}  \\
Wage (n = 3000)                          & 2.161 (0.128) & 16.016 (0.955) & 0.122 (0.011)  & \textbf{0.084 (0.011)}  \\
Wine quality (n = 4898)                  & 0.946 (0.077) & 20.402 (1.200) & 0.073 (0.010)  & \textbf{0.056 (0.009)}  \\
Yacht hydrodynamics (n = 308)            & 8.244 (0.559) & 6.244 (0.755)  & 0.323 (0.032)  & \textbf{0.245 (0.033)}  \\
Slice Localization (n = 53500)           & -             & \textbf{1.640 (5.937)}  & 4.660 (18.954) & 4.510 (17.533) \\

\hline 
\end{tabular}%
\end{table}

\begin{figure}[h]
    \centering
    \includegraphics[width=1\linewidth]{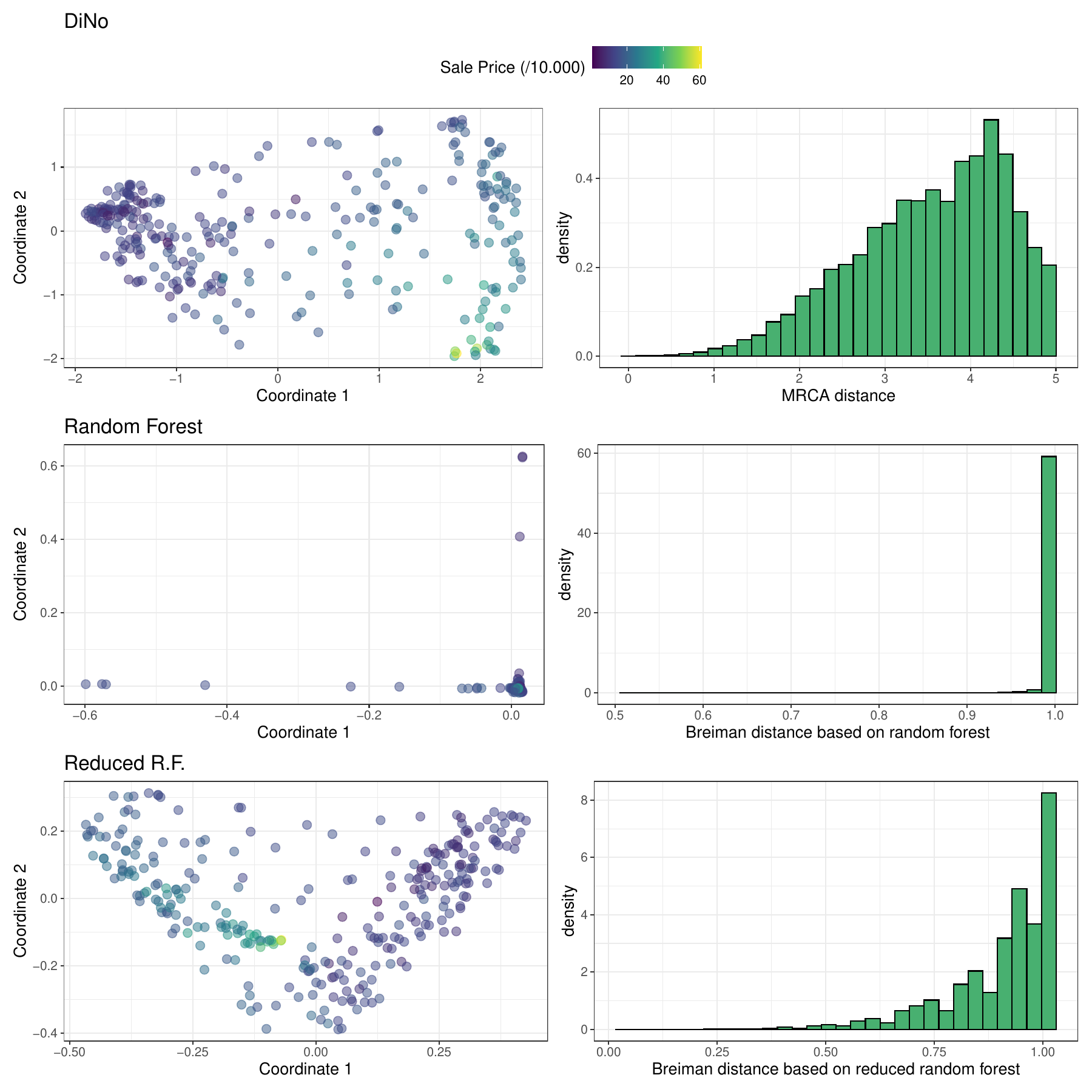}
\caption{Multidimensional scaling (MDS) plots and pairwise distance distributions for a subset of the \emph{Ames Housing} data. 
Compared with Breiman proximities from a full or reduced random forest, \DiNo{} generates embeddings with clearer structure and a more dispersed distance distribution. 
This greater spread suggests stronger discriminatory ability, making the distance attractive for clustering and other similarity–based tasks beyond prediction.}
\label{fig:ames:distances}
\end{figure}

\newpage

\clearpage
\appendix

\section*{Proof that MRCA-Based Distance is a Metric}
\label{section:appendix:proofmrca}

Consider a rooted binary tree $T$ whose terminal nodes (leaves) are the objects of interest.  
For leaves $n_1,n_2$, denote by $\operatorname{mrca}(n_1,n_2)$ their most–recent common ancestor. For any two nodes $a,b \in T$, denote by $\#(a,b)$ the number of edges on the unique path between $a$ and $b$. Note that $\#(a, b) = 0 \iff a = b$.

\textbf{MRCA-based distance}
\[
d_T(n_1,n_2)\;=\;\max\Bigl\{\#\bigl(n_1,\operatorname{mrca}(n_1,n_2)\bigr),
                       \#\bigl(n_2,\operatorname{mrca}(n_1,n_2)\bigr)\Bigr\}.
\]

\begin{proof}
We verify the four metric axioms for $d_T$.

\medskip
\noindent\textbf{1. Non-negativity.}
For any nodes $n_1,n_2$, $\#(n_1,n_2)\ge 0$, hence $d_T(n_1,n_2)\ge 0$.

\medskip
\noindent\textbf{2. Identity of indiscernibles.}
If $d_T(n_1,n_2)=0$, both terms in the maximum are $0$, so
$n_1=\operatorname{mrca}(n_1,n_2)=n_2$.
Conversely, $d_T(n_1,n_1)=\max\{0,0\}=0$.

\medskip
\noindent\textbf{3. Symmetry.}
The definition is symmetric in its arguments, since $\operatorname{mrca}(n_1, n_2) = \operatorname{mrca}(n_2, n_1)$, hence $d_T(n_1,n_2)=d_T(n_2,n_1)$.

\medskip
\noindent\textbf{4. Triangle inequality.}
Fix leaves $a,b,c$. Any binary tree on $\{a,b,c\}$ realizes exactly one of the three merge orders:
\((a,b)\) before \(c\); \((b,c)\) before \(a\); or \((a,c)\) before \(b\). Thus, we prove (4) considering these cases. First, we state the following lemma.

\paragraph{Path–decomposition lemma (via MRCA).}
Let $u,v,w$ be leaves. If $\operatorname{mrca}(u,v)$ lies on the unique path from $u$ to $\operatorname{mrca}(u,w)$, then
\[
\#\big(u,\operatorname{mrca}(u,w)\big)
=
\#\big(u,\operatorname{mrca}(u,v)\big)
+
\#\big(\operatorname{mrca}(u,v),\operatorname{mrca}(u,w)\big).
\]
This follows from uniqueness of paths in trees and additivity of edge counts.

The following inequality follows form the lemma.

\paragraph{Ancestor–path monotonicity (MRCA corollary).}
If $p$ lies on the unique path from $t$ to $q$, then
\[
\#(p,q) \le \#(t,q).
\]
Indeed, by the lemma,
$\#(t,q)=\#(t,p)+\#(p,q)\ge \#(p,q)$.

\bigskip
\noindent\textbf{Case 1: \((a,b)\) merge before \(c\).}
In this case,
\[
\operatorname{mrca}(a,c)=\operatorname{mrca}(b,c)
\quad\text{and}\quad
\operatorname{mrca}(a,b)\ \text{lies on the path}\ a\to \operatorname{mrca}(a,c).
\]

\begin{center}
\begin{minipage}{0.46\textwidth}
\begin{tikzpicture}[
  level distance=12mm,
  level 1/.style={sibling distance=36mm},
  level 2/.style={sibling distance=26mm},
  treenode/.style={
    draw, rectangle, minimum width=18mm, minimum height=8mm,
    align=center, inner sep=2pt
  }
]
\node[treenode] (r) {$\operatorname{mrca}(a,c)$}
  child { node[treenode] (x) {$\operatorname{mrca}(a,b)$}
    child { node[treenode] (a) {$a$} }
    child { node[treenode] (b) {$b$} }
  }
  child { node[treenode] (c) {$c$} };
\end{tikzpicture}
\end{minipage}\hfill
\begin{minipage}{0.5\textwidth}
\[
\begin{aligned}
d_T(a,b) &= \max\{\#(a,\operatorname{mrca}(a,b)),\ \#(b,\operatorname{mrca}(a,b))\},\\
d_T(a,c) &= \max\{\#(a,\operatorname{mrca}(a,c)),\ \#(c,\operatorname{mrca}(a,c))\},\\
d_T(b,c) &= \max\{\#(b,\operatorname{mrca}(b,c)),\ \#(c,\operatorname{mrca}(b,c))\}.
\end{aligned}
\]
\end{minipage}
\end{center}

\textit{Subcase A.} If $d_T(a,c)=\#(c,\operatorname{mrca}(a,c))$, then using $\operatorname{mrca}(a,c)=\operatorname{mrca}(b,c)$,
\[
d_T(a,c)=\#\big(c,\operatorname{mrca}(a,c)\big)=\#\big(c,\operatorname{mrca}(b,c)\big)\le d_T(b,c)\le d_T(a,b)+d_T(b,c).
\]

\textit{Subcase B.} If $d_T(a,c)=\#(a,\operatorname{mrca}(a,c))$, then by the lemma (with $u=a$, $v=b$, $w=c$),
\[
\#\big(a,\operatorname{mrca}(a,c)\big)=\#\big(a,\operatorname{mrca}(a,b)\big)+\#\big(\operatorname{mrca}(a,b),\operatorname{mrca}(a,c)\big).
\]
Now $\#(a,\operatorname{mrca}(a,b))\le d_T(a,b)$, and by the corollary applied to the path from
$b$ to $\operatorname{mrca}(b,c)=\operatorname{mrca}(a,c)$ (with $t = b$, $p=\operatorname{mrca}(a,b)$, and $q=\operatorname{mrca}(a,c)$),
\[
\#\big(\operatorname{mrca}(a,b),\operatorname{mrca}(a,c)\big)\le \#\big(b,\operatorname{mrca}(b,c)\big)\le d_T(b,c).
\]
Hence
\[
d_T(a,c)=\#\big(a,\operatorname{mrca}(a,c)\big)\le d_T(a,b)+d_T(b,c).
\]

\medskip
\noindent\textbf{Case 2: \((b,c)\) merge before \(a\).}
In this case,
\[
\operatorname{mrca}(a,c)=\operatorname{mrca}(a,b)
\quad\text{and}\quad
\operatorname{mrca}(b,c)\ \text{lies on the paths from}\ b \text{ to } \operatorname{mrca}(a,c), \text{and from } c \text{ to } \operatorname{mrca}(a,c).
\]

\begin{center}
\begin{minipage}{0.46\textwidth}
\begin{tikzpicture}[
  level distance=12mm,
  level 1/.style={sibling distance=36mm},
  level 2/.style={sibling distance=26mm},
  treenode/.style={
    draw, rectangle, minimum width=18mm, minimum height=8mm,
    align=center, inner sep=2pt
  }
]
\node[treenode] (r) {$\operatorname{mrca}(a,c)$}
  child { node[treenode] (a) {$a$} }
  child { node[treenode] (x) {$\operatorname{mrca}(b,c)$}
    child { node[treenode] (b) {$b$} }
    child { node[treenode] (c) {$c$} }
  };
\end{tikzpicture}
\end{minipage}\hfill
\begin{minipage}{0.5\textwidth}
\[
\begin{aligned}
d_T(a,b) &= \max\{\#(a,\operatorname{mrca}(a,b)),\ \#(b,\operatorname{mrca}(a,b))\},\\
d_T(a,c) &= \max\{\#(a,\operatorname{mrca}(a,c)),\ \#(c,\operatorname{mrca}(a,c))\},\\
d_T(b,c) &= \max\{\#(b,\operatorname{mrca}(b,c)),\ \#(c,\operatorname{mrca}(b,c))\}.
\end{aligned}
\]
\end{minipage}
\end{center}

\textit{Subcase A.} If $d_T(a,c)=\#(c,\operatorname{mrca}(a,c))$, decompose through $\operatorname{mrca}(b,c)$:
\[
\#\big(c,\operatorname{mrca}(a,c)\big)=\#\big(c,\operatorname{mrca}(b,c)\big)+\#\big(\operatorname{mrca}(b,c),\operatorname{mrca}(a,c)\big)\le d_T(b,c)+d_T(a,b).
\]

\textit{Subcase B.} If $d_T(a,c)=\#(a,\operatorname{mrca}(a,c))$, use that $\operatorname{mrca}(a,c)=\operatorname{mrca}(a,b)$ in this merge order:
\[
d_T(a,c)=\#\big(a,\operatorname{mrca}(a,c)\big)=\#\big(a,\operatorname{mrca}(a,b)\big)\le d_T(a,b)\le d_T(a,b)+d_T(b,c).
\]
Thus the triangle inequality holds in Case~2.

\medskip
\noindent\textbf{Case 3: \((a,c)\) merge before \(b\).}
Equivalently,
\[
\operatorname{mrca}(a,b)=\operatorname{mrca}(b,c)
\quad\text{and}\quad
\operatorname{mrca}(a,c)\ \text{lies on the paths from }\ a \text{ to } \operatorname{mrca}(a,b), \text{ and from } c \text{ to } \operatorname{mrca}(b,c).
\]

\begin{center}
\begin{minipage}{0.46\textwidth}
\begin{tikzpicture}[
  level distance=12mm,
  level 1/.style={sibling distance=36mm},
  level 2/.style={sibling distance=26mm},
  treenode/.style={
    draw, rectangle, minimum width=18mm, minimum height=8mm,
    align=center, inner sep=2pt
  }
]
\node[treenode] (r) {$\operatorname{mrca}(a,b)$}
  child { node[treenode] (b) {$b$} }
  child { node[treenode] (x) {$\operatorname{mrca}(a,c)$}
    child { node[treenode] (a) {$a$} }
    child { node[treenode] (c) {$c$} }
  };
\end{tikzpicture}
\end{minipage}\hfill
\begin{minipage}{0.5\textwidth}
\[
\begin{aligned}
d_T(a,b) &= \max\{\#(a,\operatorname{mrca}(a,b)),\ \#(b,\operatorname{mrca}(a,b))\},\\
d_T(a,c) &= \max\{\#(a,\operatorname{mrca}(a,c)),\ \#(c,\operatorname{mrca}(a,c))\},\\
d_T(b,c) &= \max\{\#(b,\operatorname{mrca}(b,c)),\ \#(c,\operatorname{mrca}(b,c))\}.
\end{aligned}
\]
\end{minipage}
\end{center}

\textit{Subcase A.} If $d_T(a,c)=\#(a,\operatorname{mrca}(a,c))$, then $\operatorname{mrca}(a,c)$ lies on the unique path from $a$ to $\operatorname{mrca}(a,b)$,
so by the ancestor–path monotonicity,
\[
d_T(a,c)=\#(a,\operatorname{mrca}(a,c))\le \#(a,\operatorname{mrca}(a,b))\le d_T(a,b)\le d_T(a,b)+d_T(b,c).
\]

\textit{Subcase B.} If $d_T(a,c)=\#(c,\operatorname{mrca}(a,c))$, then $\operatorname{mrca}(a,c)$ lies on the unique path from $c$ to $\operatorname{mrca}(b,c)$,
hence
\[
d_T(a,c)=\#(c,\operatorname{mrca}(a,c))\le \#(c,\operatorname{mrca}(b,c))\le d_T(b,c)\le d_T(a,b)+d_T(b,c).
\]

Thus the triangle inequality holds in Case~3 as well, which completes the proof.

\end{proof}

\end{document}